# Identification of Pleonastic *It* Using the Web

**Yifan Li** YIFAN@ECE.UALBERTA.CA
**Petr Musilek** MUSILEK@ECE.UALBERTA.CA
**Marek Reformat** REFORM@ECE.UALBERTA.CA
**Loren Wyard-Scott** WYARD@ECE.UALBERTA.CA
*Department of Electrical and Computer Engineering*
*University of Alberta*
*Edmonton, AB T6G 2V4 Canada*

## Abstract

In a significant minority of cases, certain pronouns, especially the pronoun *it*, can be used without referring to any specific entity. This phenomenon of pleonastic pronoun usage poses serious problems for systems aiming at even a shallow understanding of natural language texts. In this paper, a novel approach is proposed to identify such uses of *it*: the extrapositional cases are identified using a series of queries against the web, and the cleft cases are identified using a simple set of syntactic rules. The system is evaluated with four sets of news articles containing 679 extrapositional cases as well as 78 cleft constructs. The identification results are comparable to those obtained by human efforts.

## 1. Introduction

Anaphora resolution, which associates a word or phrase (the anaphor) with a previously mentioned entity (the antecedent), is an active field of Natural Language Processing (NLP) research. It has an important role in many applications where a non-trivial level of understanding of natural language texts is desired, most notably in information extraction and machine translation. To illustrate, an information extraction system trying to keep track of corporate activities may find itself dealing with news such as '*Microsoft today announced that it is adopting XML as the default file format for the next major version of its Microsoft Office software* ... ' It would be impossible to provide any insight into what Microsoft's intention is without associating the pronominal anaphors *it* and *its* with their antecedent, *Microsoft*.

Adding to the already complex problem of finding the correct antecedent, pronouns are not always used in the same fashion as shown in the earlier example. It is well-known that some pronouns, especially *it*, can occur without referring to a nominal antecedent, or any antecedent at all. Pronouns used without an antecedent, often referred to as being pleonastic or structural, pose a serious problem for anaphora resolution systems. Many anaphora resolution systems underestimate the issue and choose not to implement a specific module to handle pleonastic pronouns but instead have their input 'sanitized' manually to exclude such cases. However, the high frequency of pronoun usage in general and pleonastic cases in particular warrants that the phenomenon deserves more serious treatment. The pronoun *it*, which accounts for most of the pleonastic pronoun usages, is by far the most frequently used of all pronouns in the British National Corpus (BNC). In the Wall Street Journal Corpus (WSJ; Marcus, Marcinkiewicz, & Santorini, 1993), upon which this study is based, *it* accounts for more than 30% of personal pronoun usage. The percentage of cases where *it* lacks a nominal antecedent is also significant: previous studies have reported figures between 16% and 50% (Gundel, Hedberg, & Zacharski, 2005) while our own analysis based upon the WSJ corpus results in a value around 25%, more than half of which are pleonastic cases.

Applying criteria similar to those established by Gundel et al. (2005), the usage of *it* can be generally categorized as follows. The instances of *it* being analyzed are shown in *italics*; and the corresponding antecedents, extraposed clauses, and clefted constituents are marked by underlining.

1. Referential with nominal antecedent

   [0006:002][1] The thrift holding company said *it* expects to obtain regulatory approval and complete the transaction by year-end.

   where *it* refers to the thrift holding company.

2. Referential with clause antecedent

   [0041:029] He was on the board of an insurance company with financial problems, but he insists he made no secret of *it*.

   where *it* refers to the fact that the person was on the board of an insurance company.

   [0102:002-003] Everyone agrees that most of the nation's old bridges need to be repaired or replaced. But there's disagreement over how to *do it*.

   where *it*, together with *do*, refers to the action of repairing or replacing the bridge.

3. No antecedent – Pleonastic

   (a) Extraposition

   [0034:020] But *it* doesn't take much to get burned.

   where the infinitive clause to get burned is extraposed and its original position filled with an expletive *it*. The equivalent non-extraposed sentence is '*But to get burned doesn't take much.*'

   [0037:034] *It*'s a shame their meeting never took place.

   The equivalent non-extraposed sentence is '*That their meeting never took place is a shame.*'

   (b) Cleft[2]

   [0044:026] And most disturbing, *it* is educators, not students, who are blamed for much of the wrongdoing.

   The equivalent non-cleft version is '*And most disturbing, educators, not students, are blamed for much of the wrongdoing.*'

[1] All example sentences are selected from the WSJ corpus, with locations encoded in the format [article:sentence].

[2] Some claim that cleft pronouns should not be classified as expletive (Gundel, 1977; Hedberg, 2000). Nevertheless, this does not change the fact that the pronouns do not have nominal antecedents; hence clefts are included in this analysis.

[0591:021] *It* is partly for this reason that the exchange last week began trading in its own stock "basket" product ...

The equivalent non-cleft version is '*The exchange last week began trading in its own stock basket product partly for this reason.*'

(c) Local Situation

[0207:037] *It* was not an unpleasant evening ...

This category consists of *it* instances related to weather, time, distance, and other information about the local situation. Since the texts reviewed in this study lack instances of other subtypes, only weather and time cases are discussed.

4. Idiomatic

[0010:010] The governor couldn't make *it*, so the lieutenant governor welcomed the special guests.

This paper focuses on pleonastic cases (the third category), where each subclass carries its unique syntactic and/or semantic signatures. The idiomatic category, while consisting of non-anaphoric cases as well, is less coherent and its identification is much more subjective in nature, making it a less attractive target.

This paper is organized as follows: Section 2 provides a brief survey of related work toward both classification of *it* and identification of pleonastic *it*; Section 3 proposes a web-based approach for identification of pleonastic *it*; Section 4 demonstrates the proposed method with a case study; Section 5 follows with evaluation; and finally, Section 6 discusses the findings and presents ideas for future work.

## 2. Previous Work

As Evans (2001) pointed out, usage of *it* is covered in most serious surveys of English grammar, some of which (e.g. Sinclair, 1995) also provide classifications based on semantic categories. In a recent study, Gundel et al. (2005) classify third-person personal pronouns into the following comprehensive hierarchy:

- Noun phrase (NP) antecedent
- Inferrable
- Non-NP antecedent
  - Fact
  - Proposition
  - Activity
  - Event
  - Situation
  - Reason
- Pleonastic
  - Full extraposition
  - Full cleft
  - Truncated cleft
  - Truncated extraposition
  - Atmospheric
  - Other pleonastic
- Idiom
- Exophoric
- Indeterminate

Without going into the details of each category, it is apparent from the length of the list that the phenomenon of pleonastic *it*, and more generally pronouns without explicit nominal antecedents, have been painstakingly studied by linguists. However, despite being identified as one of the open issues of anaphora resolution (Mitkov, 2001), work on automatic identification of pleonastic *it* is relatively scarce. To date, existing studies in the area fall into one of two categories: one wherein a rule-based approach is used, and the other using a machine-learning approach.

### 2.1 Rule-based Approaches

Paice and Husk (1987) together with Lappin and Leass (1994) provide examples of rule-based systems that make use of predefined syntactic patterns and word lists. The Paice and Husk approach employs bracketing patterns such as *it ... to* and *it ... who* to meet the syntactic restrictions of extraposition and cleft. The matched portions of sentences are then evaluated by further rules represented by word lists. For example, the *it ... to* rule prescribes that one of the 'task status' words, such as *good* or *bad*, must be present amid the construct. In order to reduce false positives, general restrictions are applied on sentence features such as construct length and intervening punctuation.

Lappin and Leass's (1994) approach employs a set of more detailed rules such as *It is* `Modaladj` *that* `S` and *It is* `Cogv-ed` *that* `S`, where `Modaladj` and `Cogv` are predefined lists of modal adjectives (e.g. *good* and *useful*) and cognitive verbs (e.g. *think* and *believe*), respectively. Compared to Paice and Husk's (1987) approach, this method is much more restrictive, especially in its rigidly-specified grammatical constraints. For example, it is not clear from the original Lappin and Leass paper whether the system would be able to recognize sentences such as [0146:014] '*It isn't clear, however, whether* ... ' despite its claim that the system takes syntactic variants into consideration.

Lappin and Leass's (1994) approach is part of a larger system, and no evaluation is provided. The Paice and Husk (1987) approach, on the other hand, evaluates impressively. It has an accuracy of 93.9% in determining pleonastic constructs on the same data used for rule development, without using part-of-speech tagging or parsing.

Both rule-based systems rely on patterns to represent syntactic constraints and word lists to represent semantic constraints. This makes them relatively easy to implement and maintain. However, these features also make them less scalable – when challenged with large and unfamiliar corpora, their accuracies deteriorate. For example, Paice and Husk (1987) noticed nearly a 10% decrease in accuracy when rules developed using one subset of the corpus are applied to another subset without modifications. Boyd, Gegg-Harrison, and Byron (2005) also observed a significant performance penalty when the approach was applied to a different corpus. In other words, rule-based systems can only be as good as they are designed to be. Denber (1998) suggested using WordNet (Fellbaum, 1998) to extend the word lists, but it is doubtful how helpful this would be considering the enormous number of possible words that are not included in existing lists and the number of inapplicable words that will be identified by such an approach.

### 2.2 Machine-learning Approaches

Recent years have seen a shift toward machine-learning approaches, which shed new light on the issue. Studies by Evans (2001, 2000) and Boyd et al. (2005) are examples of this class. Both systems employ memory-based learning on grammatical feature vectors; Boyd et al.'s approach also includes a decision tree algorithm that produces less ideal results. In his attempt to place uses of *it* into seven categories, including pleonastic and nominal anaphoric among others, Evans uses 35 features to encode information such as position/proximity, lemmas, and part-of-speech, related to both the pronoun and other components of interest, such as words and noun phrases, in the sentence. Evans reported 73.38% precision and 69.25% recall for binary classification of pleonastic cases, and an overall binary classification accuracy of 71.48%. In a later study featuring MARS[3], a fully automatic pronoun resolution system that employs the same approach, Mitkov, Evans, and Orasan (2002) reported a significantly higher binary classification accuracy of 85.54% when the approach is applied to technical manuals.

Boyd et al.'s (2005) approach targets pleonastic *it* alone. It uses 25 features, most of which concern lengths of specific syntactic structures; also included are part-of-speech information and lemmas of verbs. The study reports an overall precision of 82% and recall of 71%, and, more specifically, recalls on extrapositional and cleft constructs of 81% and 45%, respectively.

In addition, Clemente, Torisawa, and Satou (2004) used support vector machines with a feature-set similar to that proposed by Evans (2001) to analyze biological and medical texts, and reported an overall accuracy of 92.7% – higher than that of their own memory-based learning implementation. Ng and Cardie (2002) built a decision tree for binary anaphoricity classification on all types of noun phrases (including pronouns) using the C4.5 induction algorithm. Ng and Cardie reported overall accuracies of 86.1% and 84.0% on the MUC-6 and MUC-7 data sets. Categorical results, however, are not reported and it is not possible to determine the system's performance on pronouns. Using automatically induced rules, Müller (2006) reported an overall accuracy of 79.6% when detecting non-referential *it* in spoken dialogs. An inter-annotator agreement study conducted in the same paper indicates that it is difficult even for humans to classify instances of *it* in spoken dialogs. This finding is supported by our own experiences.

Machine-learning approaches are able to partly circumvent the restrictions imposed by fixed word lists or rigid grammatical patterns through learning. However, their advantage also comes with a price – training is required in the initial development phase and for different corpora re-training is preferable since lemmas are part of the feature sets. Since the existing approaches fall within the area of supervised learning (i.e. training data need to be manually classified), the limited number of lemmas they gather from training may lead to degraded performance in unfamiliar circumstances. Moreover, the features used during learning are unable to reliably capture the subtleties of the original sentences, especially when considering non-technical documents. For example, the quantitative features frequently used in machine-learning approaches, such as position and distance, become less reliable when sentences contain a large number of adjuncts. Additionally, the meanings of lemmas are often domain-dependent and can vary with their local structural and lexical

[3] Available online at http://clg.wlv.ac.uk/demos/MARS/

environment – such nuances cannot be captured by the lemma features alone. In short, while machine-learning approaches generally deliver better performance classifying *it* than their rule-based counterparts do, they have their own inherent problems.

## 3. A Web Based Approach

Both syntactic patterns and semantics of various clause constituents play important roles in determining if a third-person personal pronoun is pleonastic. The role of grammar is quite obvious since both extrapositions and clefts must follow the grammatical patterns by which they are defined. For example, the most commonly seen type of *it*-extraposition follows the pattern:

| | *it* | + copula | + status | + subordinate clause |
|---|---|---|---|---|
| [0089:017] | *It* ␣ | is ␣ | easy ␣ | to see why the ancient art is on the ropes. |

In contrast, the role semantics plays here is a little obscure until one sits down and starts to "dream up exceptions" (Paice & Husk, 1987) analogous to [0074:005] ' ... *it has taken measures to continue shipments during the work stoppage.*' vis-à-vis [0367:044] ' ... *it didn't take a rocket scientist to change a road bike into a mountain bike* ... ', where referential and pleonastic cases share the same syntactic structure. Despite its less overt role, failure to process semantic information can result in a severe degradation of performance. This observation is supported by the word-list-based systems' dramatic decay in accuracy when they are confronted with text other than that they obtained their word lists from.

Like every other classification system, the proposed system strives to cover as many cases as possible and at the same time perform classification as accurately as possible. To achieve this, it attempts to make good use of both syntactic and semantic information embedded in sentences. A set of relaxed yet highly relevant syntactic patterns is first applied to the input text to filter out the syntactically inviable cases. Unlike the matching routines of some previous approaches, this process avoids detailed specification of syntactic patterns. Instead, it tries to include every piece of text containing a construct of possible interest. Different levels of semantic examinations are performed for each subtype of pleonastic constructs. For reasons discussed later in Section 3.2.2, semantic analysis is not performed on clefts. A WordNet-based analysis is used to identify weather/time cases because among the samples examined during the system's development stage, cases pertaining to this class are relatively uniform in their manner of expression. For the most complex and populous class, the extrapositions, candidates are subjected to a series of tests performed as queries against the web. Results of the queries provide direct evidence of how a specific configuration of clause constituents is generally used.

The reason that such a corpus-based approach is chosen versus applying manually constructed knowledge sources, such as a word list or WordNet, is fourfold:

1. Manually constructed knowledge sources, regardless of how comprehensive they are, contain only a small portion of general world knowledge. In the particular settings of this study, general world knowledge is used for making judgements such as which words are allowed to serve as the matrix verb of an extraposition, and even more subtle, which specific sense of a word is permitted.

2. Manually compiled knowledge sources are subject to specific manners of organization that may not satisfy the system's needs. Taking WordNet as an example, it identifies a large number of various relationships among entities, but the information is mainly organized along the axes of synonyms, hypernyms (kind-of relationship), and holonyms (part-of relationship) etc., while it is the surroundings of a particular word that are of more interest to this study.

3. Natural languages are evolving quickly. Taking English as an example, each year new words are incorporated into the language[4] and the rules of grammar have not been immune to changes either. Using a large and frequently-updated corpus such as the web allows the system to automatically adapt to changes in language.

4. Most importantly, corpora collect empirical evidence of language usage. When the sample size is large enough, as in the case of the web, statistics on how a specific construct is generally used in corpora can be employed as an indicator of its speaker's intention.

The proposed approach is also inspired by Hearst's (1992) work on mining semantic relationships using text patterns, and many other quests that followed in the same direction (Berland & Charniak, 1999; Poesio, Ishikawa, im Walde, & Vieira, 2002; Markert, Nissim, & Modjeska, 2003; Cimiano, Schmidt-Thieme, Pivk, & Staab, 2005). Unlike these investigations that focus on the semantic relationship among noun phrases, the pleonastic pronoun identification problem mandates more complex queries to be built according to the original sentences. However, the binary nature of the problem also makes it simpler to apply comparative analysis on results of multiple queries, which, in turn, leads to better immunity to noise.

Figure 1 illustrates the general work flow of the proposed system. A sentence is first preprocessed to obtain a dependency tree with part-of-speech tags, which is then passed on to the syntactic filtering component to determine whether minimum grammatical requirements of the pleonastic constructs are met. It is also during the syntactic filtering process that clefts and weather/time expressions are identified using syntactic cues and the WordNet respectively. The candidate extrapositions are thereafter used to instantiate various queries on search engines; the results returned from the queries serve as parameters for the final decision-making mechanism.

### 3.1 Preprocessing

The preprocessing component transforms the syntactic information embedded in natural language texts into machine-understandable structures. During the preprocessing stage, each word is assigned a part-of-speech tag, and the whole sentence is parsed using a dependency grammar (DG) parser. For simplicity's sake, the current system is designed to use the WSJ corpus, which is already tagged and parsed with context-free grammar (CFG). A head percolation table similar to that proposed by Collins (1999) is used to obtain the head component of each phrase. The rest of the phrase constituents are then rearranged under the head component to form the dependency tree using a procedure detailed by Xia and Palmer (2001). Figure 2 illustrates the syntactic structure of a sentence in the WSJ corpus. Both the original CFG parse tree and the derived dependency structure are shown side-by-side. Head entities are underlined in the CFG diagram and circled in the DG diagram.

[4] Metcalf and Barnhart (1999) have compiled a chronicle of many important additions to the vocabulary of American English.

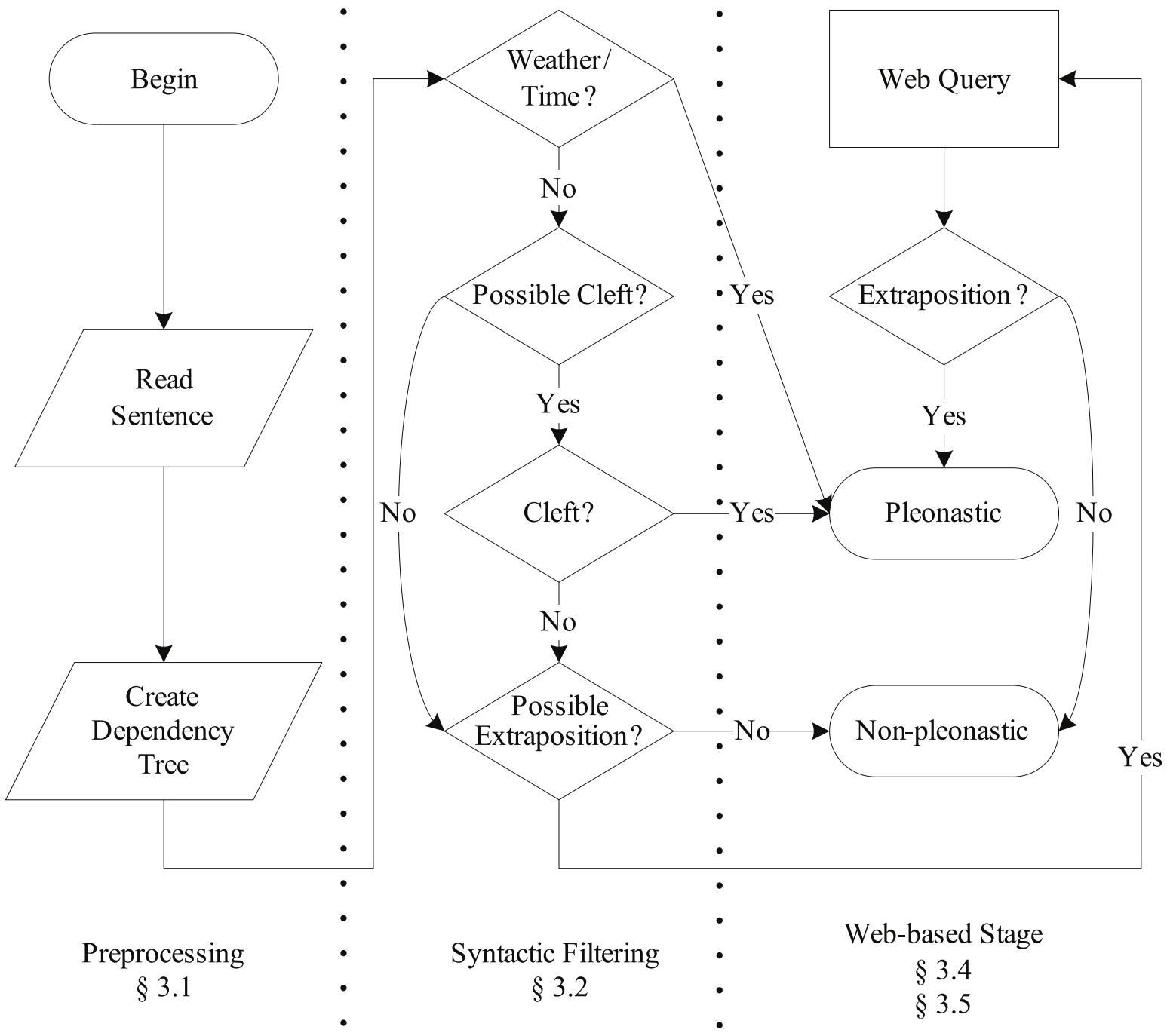

Figure 1: Illustration of the system work flow broken into three processing stages – preprocessing, syntactic filtering, and web-based analysis.

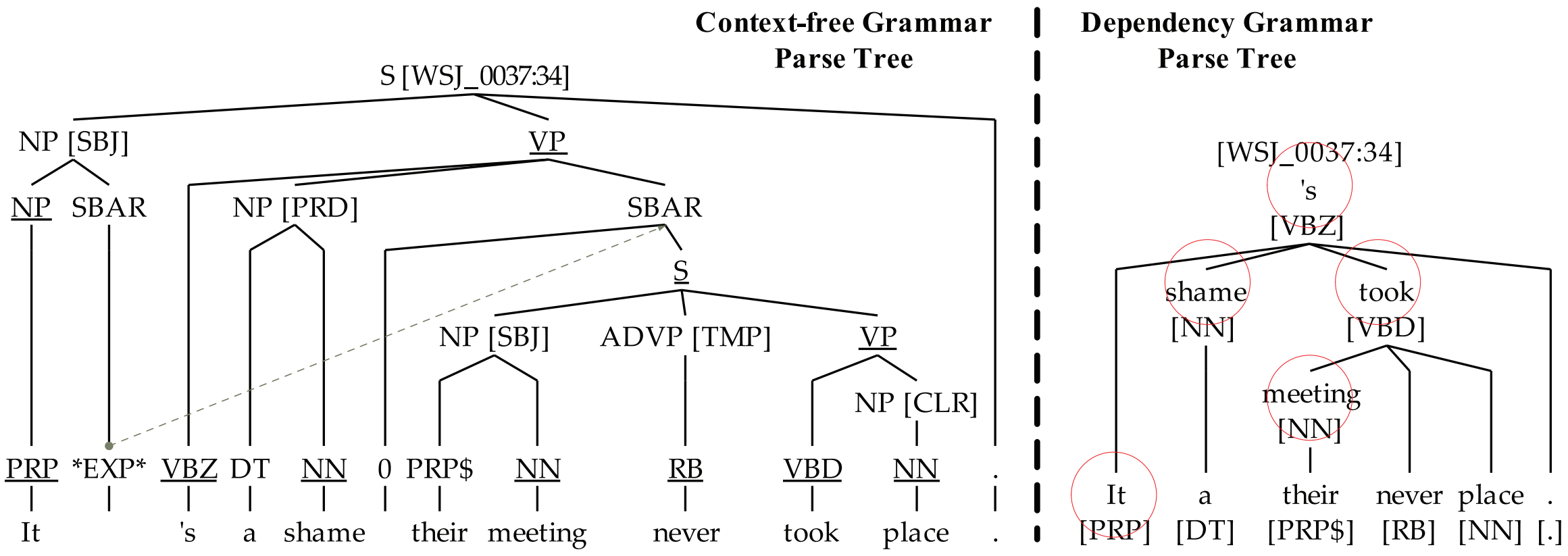

Figure 2: Illustration of a sentence's syntactic structure, both as annotated in the WSJ corpus (left) and after head percolation (right).

As shown in Figure 2, the function tags (e.g. `SBJ`, `TMP`, and `CLR`) and tracing information present in the context-free parse tree are not ported to the dependency tree. This is because real-world parsers usually do not produce such tags. Except this deliberate omission, both parse trees contain essentially the same information, only presented in different manners. In this study, dependency structure is preferred over the more popular phrase structure mainly because of its explicit marking of both the head components and the complementing/modifying relationships among various components. This feature is very helpful for instantiating the search-engine queries.

### 3.2 Syntactic Filtering

The syntactic filtering process determines whether a clause meets the grammatical requirements of an extraposition or cleft construct by matching the clause against their respective syntactic patterns.

#### 3.2.1 Extrapositions

*It*-extrapositions occur when a clause is dislocated out of its ordinary position and replaced with *it*. An *it*-extraposition usually follows the pattern:

$$\overbrace{it_{subject} + \underbrace{\left\{\begin{array}{l} be + \left\{\begin{array}{l}\texttt{noun phrase}\\ \texttt{adjective phrase}\\ \texttt{prepositional phrase}\end{array}\right\} \\ \\ \texttt{verb phrase}\end{array}\right\}}_{\texttt{matrix verb phrase}}}^{\texttt{matrix clause}} + \texttt{extraposed clause} \tag{1}$$

This pattern summarizes the general characteristics of subject *it*-extrapositions, where the pronoun *it* assumes the subject position. When the matrix verb (the verb following *it*) is the main copula *to be*, which serves to equate or associate the subject and an ensuing logical predicate, it must be followed by either a noun phrase, an adjective phrase, or a prepositional phrase.[5] There is no special requirement for the matrix verb phrase otherwise. Similarly, there is almost no restriction placed upon the extraposed clause except that a full clause should either be introduced without a complementizer (e.g. [0037:034] '*It*'s a shame their meeting never took place.') or led by *that*, *whether*, *if*, or one of the *wh*-adverbs (e.g. *how*, *why*, *when*, etc.). These constraints are developed by generalizing a small portion of the WSJ corpus and are largely in accordance with the patterns identified by Kaltenböck (2005). Compared to the patterns proposed by Paice and Husk (1987), which also cover cases such as *it ... to*, *it ... that* and *it ... whether*, they allow for a broader range of candidates by considering sentences that are not explicitly marked (such as [0037:034]). The above configuration covers sentences such as:

[5] Other copula verbs do not receive the same treatment. This arrangement is made to accommodate cases where verbs such as *to seem* and *to appear* are immediately followed by an extraposed clause.

[0529:009] Since the cost of transporting gas is so important to producers' ability to sell it, *it* helps to have input and access to transportation companies.

[0037:034] *It*'s a shame their meeting never took place.

[0360:036] *It* is insulting and demeaning to say that scientists "needed new crises to generate new grants and contracts ...[6]

[0336:019] *It* won't be clear for months whether the price increase will stick.

Except in the case of the last sentence, the above constructs are generally overlooked by the previous rule-based approaches identified in Section 2.1. As the last sample sentence illustrates, the plus sign (+) in the pattern serves to indicate a forthcoming component rather than suggest two immediately adjacent components.

Some common grammatical variants of the pattern are also recognized by the system, including questions (both direct and indirect), inverted sentences, and parenthetical expressions (Paice & Husk, 1987). This further expands the pattern's coverage to sentences such as:

[0772:006] I remembered how hard *it* was for an outsider to become accepted ...

[0562:015] "The sooner our vans hit the road each morning, the easier *it* is for us to fulfill that obligation."

[0239:009] Americans *it* seems have followed Malcolm Forbes's hot-air lead and taken to ballooning in a heady way.

Aside from being the subject of the matrix clause, extrapositional *it* can also appear in the object position. The system described here captures three flavors of object extraposition. The first type consists of instances of *it* followed by an object complement:

[0044:014] Mrs. Yeargin was fired and prosecuted under an unusual South Carolina law that <u>makes *it* a crime to breach test security</u>.

In this case the system inserts a virtual copula *to be* between the object *it* and the object complement (*a crime*), making the construct applicable to the pattern of subject extraposition. For example, the underlined part of the prior example translates into '*it* is a crime to breach test security'.

The other two kinds of object extraposition are relatively rare:

- Object of verb (without object complement)

  [0114:007] Speculation had *it* that the company was asking $100 million for an operation said to be losing about $20 million a year ...

- Object of preposition

  [1286:054] They should see to *it* that their kids don't play truant ...

These cases cannot be analyzed within the framework of subject extraposition and thus must be approached with a different pattern:

$$\texttt{verb + [preposition]}\ it_{object}\ \texttt{+ full clause} \quad (2)$$

[6] Neither *insulting* nor *demeaning* is in Paice and Husk's (1987) list of 'task status words' and therefore cannot activate the `it ... to` pattern.

The current system requires that the full clauses start with a complementizer *that*. This restriction, however, is included only to simplify implementation. Although in object expositions it is more common to have clauses led by *that*, full clauses without a leading complementizer are also acceptable.

According to Kaltenböck's (2005) analysis there are special cases in which noun phrases appear as an extraposed component, such as '*It's amazing the number of theologians that sided with Hitler.*' He noted that these noun phrases are semantically close to subordinate interrogative clauses and can therefore be considered a marginal case of extraposition. However, no such cases were found in the corpus during the annotation process and they are consequently excluded from this study.

#### 3.2.2 Cleft

*It*-clefts are governed by a slightly more restricted grammatical pattern. Following Hedberg (1990), *it*-clefts can be expressed as follows:

$$it_{subject} + \texttt{copula} + \texttt{clefted constituent} + \texttt{cleft clause} \quad (3)$$

The cleft clause must be finite (i.e. a full clause or a relative clause); and the clefted constituents are restricted to either noun phrases, clauses, or prepositional phrases.[7] Examples of sentences meeting these constraints include:

[0296:029] "*It*'s the total relationship that is important."

[0267:030] *It* was also in law school that Mr. O'Kicki and his first wife had the first of seven daughters.

[0121:048] "If the market goes down, I figure *it*'s paper profits I'm losing."

In addition, another non-canonical and probably even marginal case is also identified as a cleft:

[0296:037] I really do not understand how *it* is that Filipinos feel so passionately involved in this father figure that they want to dispose of and yet they need.

Text following the structure of this sample, where a *wh*-adverb immediately precedes *it*, is captured using the same syntactic pattern by appending a virtual prepositional phrase to the matrix copula (e.g. '*for this reason*'), as if the missing information has already been given.

Each of the examples above represents a possible syntactic construct of *it*-clefts. While it is difficult to tell the second and the third cases apart from their respective extrapositional counterparts, it is even more difficult to differentiate the first case from an ordinary copula sentence with a restrictive relative clause (RRC). For example, the following sentence,

[0062:012] "*It*'s precisely the kind of product that's created the municipal landfill monster," the editors wrote.

and its slightly modified version,

[0062:012′] "*It*'s this kind of product that's created the municipal landfill monster," the editors wrote.

[7] Adjective and adverb phrases are also possible but they are relatively less frequent and are excluded from this analysis.

are similar in construction. However, the latter is considered a cleft construct while the first is an RRC construct. To make things worse, and as pointed out by many (e.g. Boyd et al., 2005, p.3, example 5), sometimes it is impossible to make such a distinction without resorting to the context of the sentence.

Fortunately, in the majority of cases the syntactic features, especially those of the clefted constituent, provide some useful cues. In an *it*-cleft construct, the cleft clause does not constitute a head-modifier relationship with the clefted constituent, but instead forms an existential and exhaustive presupposition[8] (Davidse, 2000; Hedberg, 2000; Lambrecht, 2001). For example, '*I figure it's paper profits I'm losing.*' implies that in the context there is something (and only one thing) that the speaker is going to lose, and further associates '*paper profits*' with it. This significant difference in semantics often leaves visible traces on the syntactic layer, some of which, such as the applicability of proper nouns as clefted constituents, are obvious. Others are less obvious. The system utilizes the following grammatical cues when deciding if a construct is an *it*-cleft[9]:

- For the clefted constituent:
  - Proper nouns[10] or pronouns, which cannot be further modified by an RRC;
  - Common nouns without determiner, which generally refer to kinds[11];
  - Plurals, which violate number agreement;
  - Noun phrases that are grounded with demonstratives or possessives, or that are modified by RRCs, which unambiguously identify instances, making it unnecessary in most cases to employ an RRC;
  - Noun phrases grounded with the definite determiner *the*, and modified by an *of*-preposition whose object is also a noun phrase grounded with *the* or is in plural. These constructs are usually sufficient for introducing uniquely identifiable entities (through association), thus precluding the need for additional RRC modifiers. The words *kind*, *sort*, and their likes are considered exceptions of this rule;
  - Adverbial constructs that usually do not appear as complements. For example, phrases denoting location (*here*, *there* etc.) or a specific time (*today*, *yesterday* etc.), or a clause led by *when*; and
  - Full clauses, gerunds, and infinitives.
- For the subordinate clause:
  - Some constructs appear awkward to be used as an RRC. For example, one would generally avoid using sentences such as '*it is a place that is dirty*', as there are better alternatives. In the current implementation two patterns are considered inappropriate for RRCs, especially in the syntactic settings described in Equation 3: A) the subordinate verb phrase consists of only a copula verb and an adjective; and B) the subordinate verb phrase consists of no element other than the verb itself.

[8] This applies to canonical clefts, which do not include the class represented by [0267:030].

[9] A construct is considered an *it*-cleft if any of the conditions are met.

[10] There are exceptional cases where proper names are used with additional determiners and RRC modifiers, such as in '*the John who was on TV last night*', c.f. Sloat's (1969) account.

[11] The validity of this assertion is under debate (Krifka, 2003). Nevertheless, considering the particular syntactic setting in discussion, it is highly unlikely that bare noun phrases are used to denote specific instances.

- Combined:
  - When the clefted constituent is a prepositional phrase and the subordinate clause is a full clause, such as in the case of [0267:030], the construct is classified as a cleft[12].

Some of these rules are based on heuristics and may have exceptions, making them less ideal guidelines. Moreover, as mentioned earlier, there are cleft cases that cannot be told apart from RRCs by any grammatical means. However, experiments show that these rules are relatively accurate and provide appropriate coverage, at least for the WSJ corpus.

#### 3.2.3 Additional Filters

Aside from the patterns described in earlier sections, a few additional filters are installed to eliminate some semantically unfit constructs and therefore reducing the number of trips to search engines. The filtering rules are as follows:

- For a clause to be identified as a subordinate clause and subsequently processed for extraposition or cleft, the number of commas, dashes and colons between the clause and *it* should be either zero or more than one, a rule adopted from Paice and Husk's (1987) proposal.
- Except the copula *to be*, sentences with matrix verbs appearing in their perfect tense are not considered for either extraposition or cleft.
- When *it* is the subject of multiple verb phrases, the sentence is not considered for either extraposition or cleft.
- Sentences having a noun phrase matrix logical predicate together with a subordinate relative clause are not considered for extraposition.
- Sentences having both a matrix verb preceded by modal auxiliaries *could* or *would* and a subordinate clause led by *if* or a *wh*-adverb are not considered for extraposition. For example, [0013:017] ' ... *it could complete the purchase by next summer if its bid is the one approved by* ... ' is not considered for extraposition.

Except for the first, these rules are optional and can be deactivated in case they introduce false-negatives.

### 3.3 Using the Web as a Corpus

The first question regarding using the web as a corpus is whether it can be regarded as a corpus at all. As Kilgarriff and Grefenstette (2003) pointed out, following the definition of corpus-hood that 'a corpus is a collection of texts when considered as an object of language or literary study', the answer is yes. With the fundamental problem resolved, what remains is to find out whether the web can be an effective tool for NLP tasks.

[12] In case it is not a cleft, chances are that it is an extraposition. This assumption, therefore, does not affect the overall binary classification.

As a corpus, the web is far from being well-balanced or error-free. However, it has one feature in which no other corpus can be even remotely comparable – its size. No one knows exactly how big it is, but each of the major search engines already indexes billions of pages. Indeed, the web is so large that sometimes a misspelled word can yield tens of thousands of results (try the word *neglectible*). This sends out a mixed signal about using the web as a corpus: on the good side, even relatively infrequent terms yield sizable results; on the bad side, the web introduces much more noise than manually-compiled corpora do. In Markert and Nissim's (2005) recent study evaluating different knowledge sources for anaphora resolution, the web-based method achieves far higher recall ratio than those that are BNC- and WordNet-based, while at the same time yielding slightly lower precision. Similar things can be said about the web's diverse and unbalanced composition, which means that it can be used as a universal knowledge source – only if one can manage not to get overwhelmed by non-domain-specific information.

That being said, it is still very hard to overstate the benefits that the web offers. As the largest collection of electronic texts in natural language, it not only hosts a good portion of general world knowledge, but also stores this information using the very syntax that defines our language. In addition, it is devoid of the systematic noise introduced into manually-constructed knowledge sources during the compilation process (e.g. failure to include less frequent items or inflexible ways of information organization). Overall, the web is a statistically reliable instrument for analyzing various semantic relationships stored in natural languages by means of examples.

As also suggested by Kilgarriff (2007) and many others, it is technically more difficult to exploit the web than to use a local corpus and it can often be dangerous to rely solely on statistics provided by commercial search engines. This is mainly due to the fact that commercial search engines are not designed for corpus research. Worse, some of their design goals even impede such uses. For example, search engines skew the order of results using a number of different factors in order to provide users with the 'best' results. Combined with this is the fact that they only return results up to certain thresholds, making it essentially impossible to get unbiased results. Other annoyances include unreliable result counts, lack of advanced search features[13], and unwillingness to provide unrestricted access to their APIs. Before a new search engine specifically designed for corpus research is available, it seems we will have to work around some of those restrictions and live with the rest.

### 3.4 Design of Search Engine Queries

As discussed in previous sections, *it*-extrapositions cannot be reliably identified using syntactic signatures alone or in combination with synthetic knowledge bases. To overcome the artificial limitations imposed by knowledge sources, the proposed system resorts to the web for the necessary semantic information.

[13] For example, the wildcard (∗) feature on Google, which could be immensely useful for query construction, no longer restricts its results to single words since 2003; Yahoo's ability to support alternate words within quoted texts is limited, while MSN does not offer that feature at all.

The system employs three sets of query patterns: the *what*-cleft, the comparative expletive test, and the missing-object construction. Each set provides a unique perspective of the sentence in question. The *what*-cleft pattern is designed to find out if the sentence under investigation has a valid *what*-cleft counterpart. Since *it*-extrapositions and *what*-clefts are syntactically compatible (as shown in Section 3.4.1) and valid readings can usually be obtained by transformations from one construct to the other, the validity of the *what*-cleft is indicative of whether or not the original sentence is extrapositional. The comparative expletive test patterns are more straightforward – they directly check whether the instance of *it* can be replaced by other entities that cannot be used expletively in the same context as that of an extrapositional *it*. If the alternate construct is invalid, the original sentence can be determined as expletive. The third set of patterns are supplemental. They are intended only for identifying the relatively rare phenomenon of missing-object construction, which may not be reliably handled by the previous pattern sets.

Designing the appropriate query patterns is the most important step in efforts to exploit large corpora as knowledge sources. For complex queries against the web, it is especially important to suppress unwanted uses of certain components, which could result from different word senses, different sentence configuration, or a speaker's imperfect command of the language. For example, the query "*it is a shame that*" could return both a valid extrapositional construct and an RRC such as '*It is a shame that is perpetuated in his life*'; and the query "*what is right is that*" could return both valid *what*-clefts and sentences such as '*Why we ought to do what is right is that ...* ' This study employs three different approaches to curb unwanted results:

- The first and most important measure is comparative analysis – pairs of similarly-constructed queries are sent out to the search engine and the ratios of result counts are used for the decision. This method is effective for problems caused by both different sentence configuration and bad language usage, since generally neither contribute a fraction of results large enough to significantly affect the ratio. The method also provides a normalized view of the web because what is of interest to this study is not exactly how frequently a specific construct is used, but whether it is more likely to carry a specific semantic meaning when it is used.

- The second measure is to use stubs in query patterns, as detailed in the following sections. Stubs help ensure that the outcomes of queries are syntactically and semantically similar to the original sentences and partly resolve the problems caused by word sense difference.

- Finally, when it is infeasible to use comparative analysis, part of the query results are validated to obtain an estimated number of valid results.

### 3.4.1 Query Pattern I: The *What*-cleft

The first query pattern,

$$\texttt{\textit{What} + verb phrase + copula + stub} \quad (4)$$

is a *what*-(pseudo-)cleft construct that encompasses matrix-level information found in an *it*-extraposition. The pattern is obtained using a three-step transformation as illustrated

below:

| | | |
|---|---|---|
| | `it + verb phrase + clause` | |
| | *It* ␣ is easy ␣ to see why the ancient art is on the ropes. [0089:017] | |
| 1) | ⇓ | |
| | `clause + verb phrase` | |
| | To see why the ancient art is on the ropes ␣ is easy. | |
| 2) | ⇓ | (5) |
| | `What + verb phrase + copula + clause` | |
| | *What* ␣ is easy ␣ is ␣ to see why the ancient art is on the ropes. | |
| 3) | ⇓ | |
| | `What + verb phrase + copula + stub` | |
| | *What* ␣ is easy ␣ is ␣ to | |

Step 1 transforms the original sentence (or clause) to the corresponding non-extraposition form by removing the pronoun *it* and restoring the information to the canonical subject-verb-complement order. In the above example, the clause *to see* ... is considered the real subject and is moved back to its canonical position. The non-extraposition form is subsequently converted during step 2 to a *what*-cleft that highlights its verb phrase. Finally, in step 3, the subordinate clause is reduced into a stub to enhance the pattern's coverage. The choice of stub depends on the structure of the original subordinate clause: *to* is used when the original subordinate clause is an infinitive, a gerund, or a *for* ... infinitive construct[14]. For the rest of the cases, the original complementizer, or *that*, in the case where there is no complementizer, is used as stub. The use of a stub in the pattern imposes a syntactic constraint, in addition to the ones prescribed by the pronoun *what* and the copula *is*, that demands a subordinate clause be present in query results. The choice of stubs also reflects, to a certain degree, the semantics of the original texts and therefore can be seen as a weak semantic constraint.

Below are a few other examples of the *what*-cleft transformation:

[0059:014] *It* remains unclear whether the bond issue will be rolled over. ⇒
*What* remains unclear is whether

[0037:034] *It*'s a shame their meeting never took place. ⇒
*What* is a shame is that

The *what*-cleft pattern only identifies whether the matrix verb phrase is capable of functioning as a constituent in an *it*-extraposition. Information in the subordinate clauses is discarded because this construct is used relatively infrequently and adding extra restrictions to the query will prohibit it from yielding results in many cases.

Some *it*-extraposition constructs such as '*it appears that* ... ' and '*it is said that* ... ' do not have a valid non-extraposition counterpart, but the *what*-cleft versions often bear certain degrees of validity and queries instantiated from the pattern will often yield results (albeit not many) from reputable sources. It is also worth noting that although the input and output constructs of the transformation are syntactically compatible, they are not necessarily equivalent in terms of givenness (whether and how information in one sentence has been entailed by previous discourse). Kaltenböck (2005) noted that the percentage of extrapositional *it* constructs carrying new information varies greatly depending on the category of the text. In contrast, a *what*-cleft generally expresses new information in the subordinate clause. The presupposed contents in the two constructs are different, too. *What*-clefts, according to Gundel (1977), from which the *it*-clefts are derived, have the same existential and exhaustive presuppositions carried by their *it*-cleft counterparts. On the other hand, the *it*-extrapositions, which are semantically identical to their corresponding non-extrapositions, lack such presuppositions or, at most, imply them at a weaker strength (Geurts & van der Sandt, 2004). These discrepancies hint that a derived *what*-cleft is a 'stronger' expression than the original extraposition, which may have been why queries instantiated from the pattern tend to yield considerably less results.

[14] According to Hamawand (2003), the *for* ... infinitive construct carries distinct semantics; reducing it to the infinitive alone changes its function. However, with only a few exceptional cases, we find this reduction generally acceptable. i.e. The lost semantics does not affect the judgment of expletiveness.

Another potential problem with this pattern is its omission of the subordinate verb, which occasionally leads to false positives. For example, it does not differentiate between '*it helps to have input and access to transportation companies*' and '*it helps expand our horizon*'. This deficiency is accommodated by additional query patterns.

#### 3.4.2 Query Pattern II: Comparative Expletiveness Test

The second group of patterns provides a simplified account of the original text in a few different flavors. After execution, the results from individual queries are compared to assess the expletiveness of the subject pronoun. This set of patterns takes the following general form:

$$\texttt{pronoun + verb phrase + simplified extraposed clause} \qquad (6)$$

The only difference among individual patterns lies in the choice of the matrix clause subject pronoun: *it*, *which*, *who*, *this*, and *he*. When the patterns are instantiated and submitted to a search engine, the number of hits obtained from the *it* version should by far outnumber that of the other versions combined if the original text is an *it*-extraposition; otherwise the number of hits should be at least comparable. This behavior reflects the expletive nature of the pronoun in an *it*-extraposition, which renders the sentence invalid when *it* is replaced with other pronouns that have no pleonastic use.

A simplified extraposed clause can take a few different forms depending on its original structure:

| **Original Structure** | **Simplified** |
|---|---|
| infinitive (*to meet you*) | infinitive + stub |
| *for* ... infinitive[15](*for him to see the document*) | infinitive + stub |
| gerund (*meeting you*) | gerund + stub |
| full clause led by complementizer (*it is a shame that their meeting never took place*) | complementizer + stub |
| full clause without complementizer (*it is a shame their meeting never took place*) | *that* + stub |

Table 1: Simplification of extraposed clause

[15] The *for* ... passive-infinitive is transformed into active voice (e.g. '*for products to be sold*'→'*to sell products*').

Similar to the case of Pattern I, the stub is used both as a syntactic constraint and a semantic cue. Depending on the type of search engine, the stub can be either *the*, which is the most widely used determiner, or a combination of various determiners, personal pronouns and possessive pronouns, all of which indicate a subsequent noun phrase. In the case that an infinitive construct involves a subordinate clause led by a *wh*-adverb or *that*, the complementizer is used as stub. This arrangement guarantees that the results returned from the query conform to the original text syntactically and semantically. A null value should be used for stubs in an object position if the original text lacks a nominal object. To illustrate the rules of transformation, consider the following sentence:

[0044:010] "My teacher said *it* was OK for me to use the notes on the test," he said.

The relevant part of the sentence is:

| | | | | |
|---|---|---|---|---|
| `it` | `+` | `verb phrase` | `+` | `clause` |
| *it* | ␣ | was OK | ␣ | for me to use the notes on the test |

Applying the clause simplification rules, the first query is obtained:

| | | | | |
|---|---|---|---|---|
| `it` | `+` | `verb phrase` | `+` | `simplified clause` |
| *it* | ␣ | was OK | ␣ | to use the |

The second query is generated by simply replacing the pronoun *it* with an alternative pronoun:

| | | | | |
|---|---|---|---|---|
| `alternative pronoun` | `+` | `verb phrase` | `+` | `simplified clause` |
| *he* | ␣ | was OK | ␣ | to use the |

Google reports 94,200 hits for the *it* query, while only one page is found using the alternative query. Since the pronoun *it* can be used in a much broader context, replacing *it* with *he* alone hardly makes a balanced comparison. Instead, the combination of *which*, *who*, *this*, and *he* is used, as illustrated in the following examples:

[0044:010] "My teacher said *it* was OK for me to use the notes on the test," he said. ⇒

$\left\{\begin{array}{l} \textit{it} \\ \textit{which/who/this/he} \end{array}\right\}$ was ok to use the

[0089:017] *It* is easy to see why the ancient art is on the ropes. ⇒

$\left\{\begin{array}{l} \textit{it} \\ \textit{which/who/this/he} \end{array}\right\}$ is easy to see why

A special set of patterns is used for object extrapositions[16] to accommodate their unique syntactic construct:

$$\texttt{verb + [preposition] pronoun + } \textit{that} \texttt{ + stub} \tag{7}$$

Stubs are chosen according to the same rules for the main pattern set, however only one alternative pronoun – *them* – is used.

[16] Instances containing object complements are treated under the framework of subject extraposition and are not included here.

[0114:007] Speculation had *it* that the company was asking $100 million for an operation said to be losing about $20 million a year ... ⇒
had $\left\{ \begin{array}{l} \textit{it} \\ \textit{them} \end{array} \right\}$ *that* the

#### 3.4.3 Query Pattern III: Missing-object Construction

One search engine annoyance is that they ignore punctuation marks. This means one can only search for text that matches a specific pattern string, but not sentences that end with a pattern string. The stubs used in Pattern II are generally helpful for excluding sentences that are semantically incompatible with the original from the search results. However, under circumstances where no stub is attached to the queries (where the query results should ideally consist of only sentences that end with the query string), the search engine may produce more results than needed. Sentences conforming to the pattern `it + copula + missing-object construction`, such as (referring to a book) '*it is easy to read*', present one such situation. What is unique about the construction – and why special treatment is needed – is that a missing-object construction usually has an *it*-extraposition counterpart in which the object is present, for example '*it is easy to read the book*'. Since the missing-object constructions are virtually the same (only shorter) as their extrapositional counterparts, there is a good chance for them to be identified as extrapositions. The following are some additional examples of the missing-object construction:

[0290:025] Where non-violent civil disobedience is the centerpiece, rather than a lawful demonstration that may only attract crime, *it* is difficult to justify.

[0018:024-025] No price for the new shares has been set. Instead, the companies will leave *it* up to the marketplace to decide.

[0111:005] He declined to elaborate, other than to say, "*It* just seemed the right thing to do at this minute.

Two sets of patterns are proposed[17] to identify the likes of the foregoing examples. The first pattern, the compound adjective test, is inspired by Nanni's (1980) study considering the *easy*-type adjective followed by an infinitive (also commonly termed *tough* construction) as a single complex adjective. The pattern takes the form

$$\texttt{stub + adjective}_{base}\texttt{-}\textit{to}\texttt{-verb} \quad (8)$$

where the stub, serving to limit the outcome of the query to noun phrases, takes a combination of determiners or *a/an* alone; the original adjective is also converted to its base form $\texttt{adjective}_{base}$ if it is in comparative or superlative form. Expanding on Nanni's original claims, the pattern can be used to evaluate all adjectives[18] as well as constructs furnished with *for* ... infinitive complements. The following example demonstrates the pattern's usage:

[17] Preliminary experiments have confirmed the effectiveness of the patterns. However, due to sparseness of samples belonging to this class, they are not included in the reported evaluation.

[18] This is based on the observation that compounds such as 'ready-to-fly' (referring to model aircrafts) exist, and that it is hard to obtain a complete enumeration of the *easy*-type adjectives.

[0258:024] The machine uses a single processor, which makes *it* easier to program than competing machines using several processors. ⇒ an easy-to-program

The second set consists of two patterns used for comparative analysis with the same general profile:

$$\textit{that} \texttt{ + verb}_{gerund} \texttt{ + stub} \tag{9}$$

where $\texttt{verb}_{gerund}$ is the gerund form of the original infinitive. The complementizer *that* is used for the sole purpose of ensuring that $\texttt{verb}_{gerund}$ appears as the subject of a subordinate clause in all sentences returned by the queries. In other words, phrases such as '*computer programming*' and '*pattern matching*' are excluded. For the first pattern, the stub is a combination of prepositions (currently *in* and *from* are chosen); for the second one, a combination of determiners or *the* alone is used. For example:

[0258:024] The machine uses a single processor, which makes *it* easier to program than competing machines using several processors. ⇒

$$\textit{that} \text{ programming } \left\{ \begin{array}{l} \textit{in|from} \\ \textit{the} \end{array} \right\}$$

This set of patterns tests the transitivity of the verb in a semantic environment similar to that of the original sentence. If the verb is used transitively more often, the pattern with determiners should yield more results, and vice versa. As supported by all preceding sample sentences, a usually-transitive verb used without an object[19] is a good indicator of missing-object construction and the sentence should be diagnosed as referential.

#### 3.4.4 Query Instantiation

Patterns must be instantiated with information found in original sentences before they are submitted to a search engine. Considering the general design principles of the system, it is not advisable to instantiate the patterns with original texts – doing so significantly reduces the queries' coverage. Instead, the object of the matrix verb phrase is truncated and the matrix verb expanded in order to obtain the desired level of coverage.

The truncation process provides different renditions based on the structure of the original object:

- Adjective phrases:
  Only the head word is used. When the head word is modified by *not* or *too*, the modifier is also retained in order to better support the *too ... to* construct and to maintain compatibility with the semantics of the original text.

- Common noun phrases:

  - with a possessive ending/pronoun, or an *of*-preposition:
    The phrase is replaced by `$PRPS$` plus the head word. `$PRPS$` is either a list of possessive pronouns or one of those more widely used, depending on caliber of the search engine used. For example, '*his location*' can be expanded to '*its* | *my* | *our* | *his* | *her* | *their* | *your location*'.

[19] An omitted object of a preposition (e.g. '*It is difficult to account for.*') has the same effect, but it is identifiable through syntactic means alone.

- with determiners:
    The phrase is replaced by a choice of `$DTA$`, `$DTTS$`, `$DTTP$`, or a combination of `$DTA$` and `$DTTS$`, plus the head word. `$DTA$` is a list of (or one of the) general determiners (i.e. *a*, *an*, *any* etc.). `$DTTS$` refers to the combination of the definite article *the* and the singular demonstratives *this* and *that*. `$DTTP$` is the plural counterpart of `$DTTS$`. The choice is based on the configuration of the original text so as to maintain semantic compatibility.
  - without determiner:
    Only the head word is used.

- Proper nouns and pronouns:
  The phrase is replaced by `$PRP$`, which is a list of (or one of the) personal pronouns.

- Prepositional phrases:
  The object of the preposition is truncated in a recursive operation.

- Numeric values:
  The phrase '*a lot*' is used instead.

Matrix verbs are expanded to include both the simple past tense and the third person singular present form with the aid of WordNet and some generic patterns. Where applicable, particles such as *out* and *up* also remain attached to the verb.

Generally speaking, truncation and expansion are good ways of boosting the patterns' coverage. However, the current procedures of truncation are still crude, especially in their handling of complex phrases. For example, the phrase '*a reckless course of action*' ([0198:011]) yields '`$PRPS$` course', which results in a total loss of the original semantics. Further enhancements of the truncation process may improve the performance but the improvement will likely be limited due to the endless possibilities of language usage and constraints imposed by search engines.

Aside from truncating and expanding the original texts, a stepped-down version of Pattern II, denoted Pattern II′, is also provided to further enhance the system's coverage. The current scheme is to simply replace the extraposed clause with a new stub – *to* – if the original extraposed clause is an infinitive, a *for* ... infinitive, or a gerund construct. For example,

[0089:017] *It* is easy to see why the ancient art is on the ropes. ⇒

$$\left\{\begin{array}{l} \textit{it} \\ \textit{which/who/this/he} \end{array}\right\} \text{ is easy to}$$

In other situations, no downgraded version is applied.

### 3.5 Binary Classification of *It*-extraposition

Five factors are taken into consideration when determining whether the sentence in question is an *it*-extraposition:

**Estimated popularity of the *what*-cleft construct (query Pattern I)** denoted as

$$W = n_w \times v_w$$

where $n_w$ is the number of results reported by the search engine, and $v_w$ is the percentage of valid instances within the first batch of snippets (usually 10, depending on the search engine service) returned with the query. Validation is performed with a case-sensitive regular expression derived from the original query. Since the *what*-cleft pattern is capitalized at the beginning, the regular expression only looks for instances appearing at the beginning of a sentence. It is particularly important to validate the results of *what*-cleft queries because some search engines can produce results based on their own interpretation of the original query. For example, Google returns pages containing "*What's found is that*" for the query "*What found is that*", which might be helpful for some but is counterproductive for the purpose of this study.

**Result of the comparative expletiveness test (query Pattern II)**

denoted as

$$r = \frac{n_X}{n_{it}}$$

where $n_{it}$ is the number of results obtained from the original *it* version of the query, and $n_X$ is the total number of results produced by replacing *it* with other pronouns such as *which* and *who*. The smaller the ratio $r$ is, the more likely that the sentence being investigated is an extraposition. Extrapositional sentences usually produce an $r$ value of 0.1 or less. When both versions of the query yield insufficient results ($max(n_{it}, n_X) < N_{min}$), $r$ takes the value $R_{scarce} = 1000$. Since *it*-extrapositions are relatively rare, it is better to assume that a sentence is not extrapositional when there is insufficient data to judge otherwise. In the case where $n_X$ is sufficient but the *it* version of the query produces no result ($n_X >= N_{min}$ AND $n_{it} = 0$), $r$ takes the value $R_{zero} = 100$. Values of $R_{zero}$ and $R_{scarce}$ are large numbers chosen arbitrarily, mainly for visualization purposes. In other words both $R_{zero}$ and $R_{scarce}$ hint that the sentence is probably not extrapositional, however neither indicates the degree of likelihood.

**Result of the stepped-down comparative expletiveness test**

denoted as $r' = \frac{n'_X}{n'_{it}}$, where $n'_{it}$ and $n'_X$ are the number of results returned from the *it* version and the alternate version of the stepped-down queries (c.f. Section 3.4.4, Page 359). The stepped-down queries are 'simplified' versions of the queries used to calculate $r$. Due to this simplification, $r'$ is usually more sensitive to extrapositions. However not all queries have stepped-down versions, in which case the original queries are reused, causing $r' = r$. Similar to the way $r$ is defined, $r'$ also takes the values $R_{scarce}$ and $R_{zero}$ in special situations.

**Synthesized expletiveness**

A new variable $R$ is defined based on the values of $r$, $n_{it}$, $n_X$ , and $r'$:

$$R = \begin{cases} r, & \text{if } \max(n_{it}, n_X) \geq N_{min}, \\ r', & \text{if } \max(n_{it}, n_X) < N_{min}. \end{cases}$$

If the original queries yield enough results, $R$ takes the value of $r$ since the original queries better preserve sentence context and are generally more accurate. However, when original queries fail, the system resorts to the back-up method of using the stepped-down queries and bases its judgement on their results instead. Overall, $R$ can be seen as a synthesized indicator of how the subject pronoun is generally used in a similar syntactic and semantic setting to that of the original sentence.

**Syntactic structure of the sentence**

denoted as $S$, a binary variable indicating if the sentence under investigation belongs to a syntactic construct that is more prone to generating false-positives. On average the *what*-cleft queries yield fewer results and are less reliable since they cannot be used to provide comparative ratios. However, they are still useful as the last line of defence to curb the impacts of certain syntactic constructs that repeatedly cause the comparative expletive tests to produce false-positives. Currently only one construct is identified – the `it verb infinitive` construct, as in '*it helps to have input from everyone*' and '*it expects to post the results tomorrow*'. Therefore,

$$S = \begin{cases} \text{TRUE}, & \text{if sentence matches } \texttt{it verb infinitive}, \\ \text{FALSE}, & \text{otherwise.} \end{cases}$$

The final binary classification of it-extraposition, $E$, is defined as follows:

$$E = \begin{cases} ((R < R_{exp}) \text{ AND } (W > N_{min})), & \text{if } S = \text{TRUE}, \\ (R < R_{exp}), & \text{if } S = \text{FALSE}. \end{cases} \tag{10}$$

where $N_{min}$ and $R_{exp}$, set to 10 and 0.15 respectively in this study, are threshold constants chosen based upon empirical observations. In other words, the system recognizes an instance of *it* as extrapositional if it is unlikely (by comparing $R$ to $R_{exp}$) that an alternative pronoun is used in its place under the same syntactic and semantic settings. For `it verb infinitive` constructs, it is also required that the sentence has a viable *what*-cleft variant (by comparing $W$ to $N_{min}$).

It is worth noting that today's major commercial search engines do not return the exact number of results for a query but rather their own estimates. The negative effect of this is somewhat mitigated by basing the final decision on ratios instead of absolute numbers.

## 4. Case Study

To better illustrate the system work flow, two sample sentences are selected from the WSJ corpus to be taken through the whole process. The first sample, [0231:015], is classified as an it-extraposition; the other, [0331:033] (with the preceding sentence providing context), is a referential case with a nominal antecedent. Some particulars of the implementation are also discussed here.

[0231:015] A fund manager at a life-insurance company said three factors make *it* difficult to read market direction.

[0331:032-033] Her recent report classifies the stock as a "hold." But *it* appears to be the sort of hold one makes while heading for the door.

### 4.1 Syntactic Filtering

First, the syntactic structures of each sentence are identified and dependencies among the constituents are established, as shown in Figures 3 and 4.

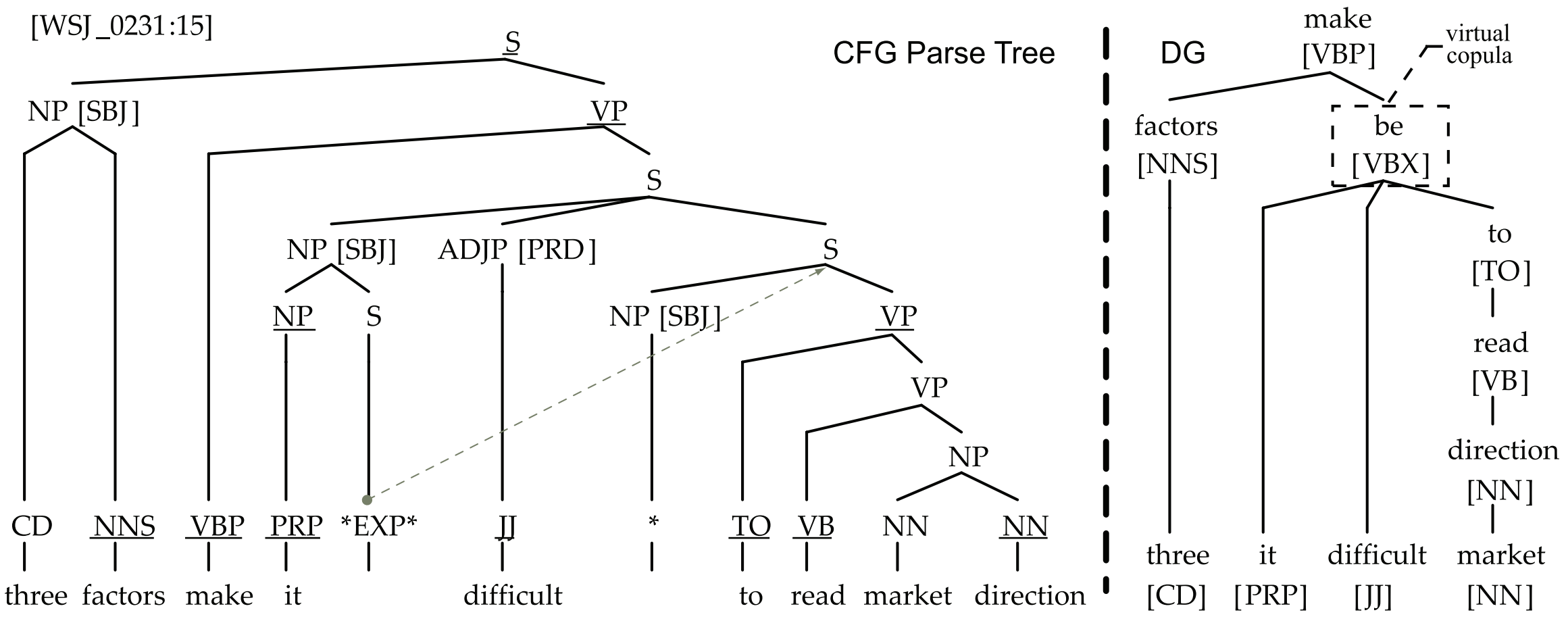

Figure 3: Syntactic structure of [0231:015] (fragment)

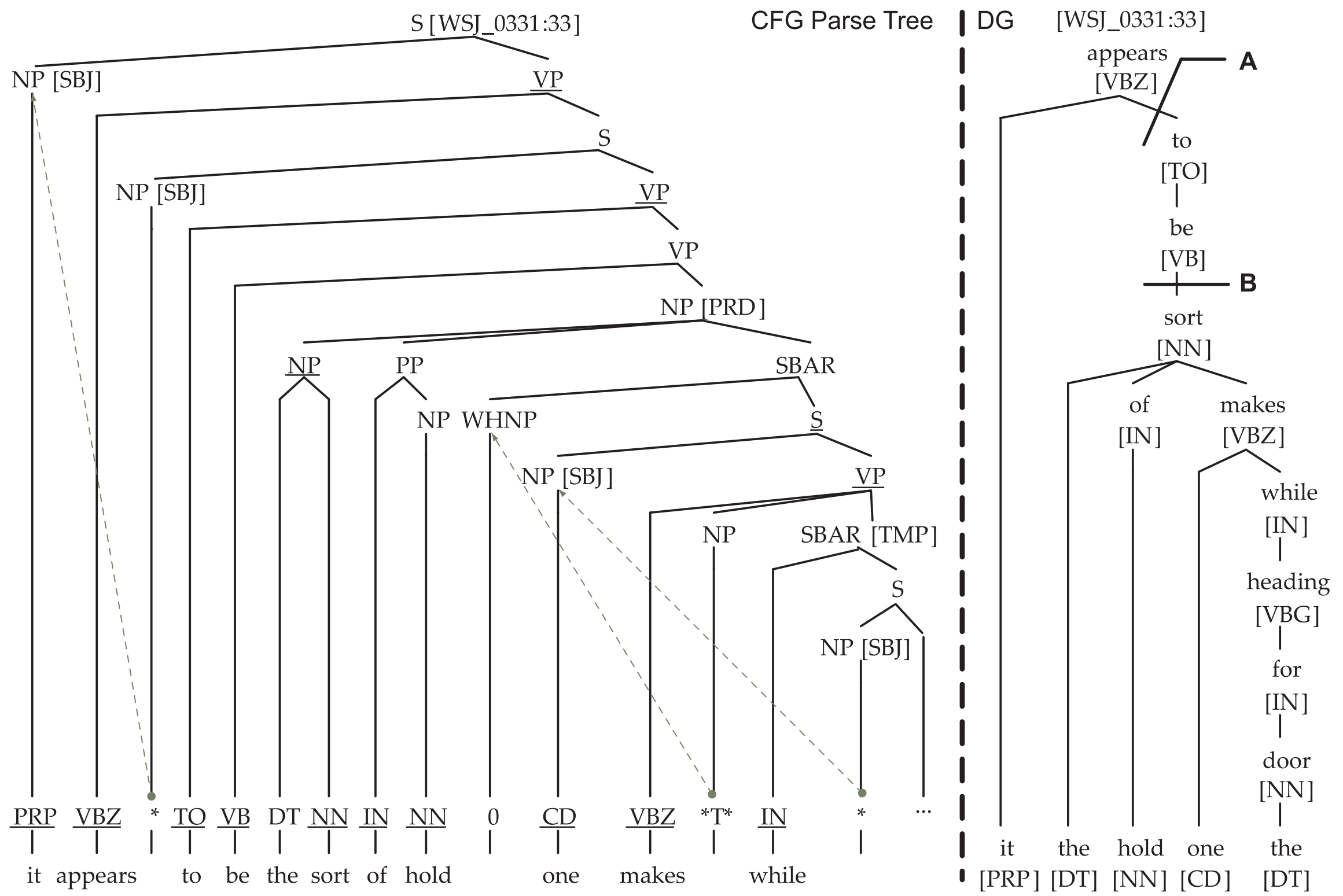

Figure 4: Syntactic structure of [0331:033] (fragment). Readings A and B, as indicated in the DG parse tree, are discussed in the text.

In sample sentence [0231:015], the expletive *it* appears as the object of the verb *makes* and is followed by the object complement *difficult*, therefore a virtual copula (tagged VBX) is created in the dependency tree in order to treat it under the same framework as subject *it*-extrapositions. For [0331:033], two different readings are produced – one by assuming *appears* to be the matrix verb (reading A, c.f. Figure 4), the other by taking *be* (reading B). This is accomplished by 'drilling' down the chain of verbs beginning with the parent verb of the *it* node. Once at the top of the chain, the system starts a recursive process to find verbs and infinitives that are directly attached to the current node and moves down to the newly found node. The process is interrupted if the current verb node is furnished with elements other than verbal or adverbial complements/modifiers.

During the filtering process, various components of the sentences are identified, as listed in Table 2.

| **Sentence** | **Reading** | **Matrix** | | **Conjunction** | **Subordinate** | | |
|---|---|---|---|---|---|---|---|
| | | **Verb** | **Object** | | **Subject** | **Verb** | **Object** |
| 0231:015 | | be | difficult | | | to read | direction |
| 0331:033 | A | appears | | | | to be | sort |
| 0331:033 | B | be | sort | THAT | One | | |

Table 2: Component breakdown of the case study samples

### 4.2 Pattern Instantiation

Using the components identified in Table 2, five queries are generated for each reading, as listed in Tables 3-5. Patterns II′-*it* and II′-others refer to the stepped-down versions (c.f. Section 3.4.4, Page 359) of II-*it* and II-others respectively. The queries shown here are generated specifically for Google and take advantage of features only available in Google. To use an alternative search engine such as Yahoo, the component expansions and determiner lists have to be turned off, and separate queries need to be prepared for individual pronouns. In order to get accurate results, the queries must be enclosed in double quotes before they are sent to search engines.

| **Pattern** | **Query** | **Results** |
|---|---|---|
| I | what is\|was\|'s difficult is\|was to | 1060 |
| II-*it* | it is\|was\|'s difficult to read the\|a\|an\|no\|this\|these\|their\|his\|our | 3960 |
| II-others | which\|this\|who\|he is\|was\|'s difficult to read the\|a\|an\|no\|this\|these\|their\|his\|our | 153 |
| II′-*it* | it is\|was\|'s difficult to | $6.3 \times 10^6$ |
| II′-others | which\|this\|who\|he is\|was\|'s difficult to | $1.5 \times 10^5$ |

Table 3: Queries for [0231:015]

| Pattern | Query | Results |
|---|---|---|
| I | what appears\|appeared is\|was to | 44 |
| II-*it* | it appears\|appeared to be the\|a\|an\|no\|this\|these\|their\|his\|our | $7.5 \times 10^4$ |
| II-others | which\|this\|who\|he appears\|appeared to be the\|a\|an\|no\|this\|these\|their\|his\|our | $3.2 \times 10^5$ |
| II′-*it* | it appears\|appeared to | $2.2 \times 10^6$ |
| II′-others | which\|this\|who\|he appears\|appeared to | $2.6 \times 10^6$ |

Table 4: Queries for [0331:033], Reading A

| Pattern | Query | Results |
|---|---|---|
| I | what is\|was\|'s its\|my\|our\|his\|her\|their\|your sort is\|was that | 0 |
| II-*it* | it is\|was\|'s its\|my\|our\|his\|her\|their\|your sort that the\|a\|an\|no\|this\|these\|they\|we\|he\|their\|his\|our | 0 |
| II-others | which\|this\|who\|he is\|was\|'s its\|my\|our\|his\|her\|their\|your sort that the\|a\|an\|no\|this\|these\|they\|we\|he\|their\|his\|our | 0 |
| II′-*it* | Same as II-*it* | 0 |
| II′-others | Same as II-others | 0 |

Table 5: Queries for [0331:033], Reading B

### 4.3 Query Results and Classification

For every reading, the number of results for each of the five queries ($n_w$ for Pattern I; $n_{it}$ for II-*it*; $n_X$ for II-others; $n'_{it}$ for II′-*it*; and $n'_X$ for II′-others) is obtained from the search engine; the first 10 results for the *what*-cleft query are also validated to obtain the estimated percentage ($v_w$) of valid constructs. $W(= n_w \times v_w)$, $r(= n_X/n_{it})$, $r'(= n'_X/n'_{it})$, and $R$ (choosing between either $r$ or $r'$ depending on whether $max(n_{it}, n_X) \geq 10$) are then calculated accordingly, as recorded in Table 6.

| Query | $n_w$ | $v_w$ | $n_{it}$ | $n_X$ | $n'_{it}$ | $n'_X$ | $W$ | $r$ | $r'$ | $R$ |
|---|---|---|---|---|---|---|---|---|---|---|
| [0231:015] | 1060 | 70% | 3960 | 153 | 6.3E6 | 1.5E5 | 742 | 0.04 | 0.02 | 0.04 |
| [0331:033].A | 44 | 0% | 7.5E4 | 3.2E5 | 2.2E6 | 2.6E6 | 0 | 4.3 | 1.2 | 4.3 |
| [0331:033].B | 0 | - | 0 | 0 | 0 | 0 | 0 | 1000 | 1000 | 1000 |

Table 6: Query results for the case study sample sentences

What appears suspicious is that $v_w$ is set to 0 for reading [0331:033].A, which means no valid instances are found. A quick look at the returned snippets reveals that, indeed, none of the 10 snippets has the queried contents at the beginning of sentence. Also note that for reading [0331:033].B, both $r$ and $r'$, and consequently $R$ have all been set to $R_{scarce} = 1000$ since no query produced enough results.

It can be decided from Table 2 that readings [0231:015] and [0331:033].B do not bear the `it verb infinitive` construct, hence $S$ = FALSE; and for [0331:033].A $S$ = TRUE. Applying Equation 10 in Section 3.5, for [0231:015] and [0331:033].B, the final classification $E$ is only based on whether $R$ is sufficiently small ($R < 0.15$). For [0331:033].A, the system also needs to check whether the *what*-cleft query returned sufficient valid results ($W > 10$). The final classifications are listed in Table 7.

| **Sentence** | **Reading** | $W$ | $S$ | $R$ | $E_{reading}$ | $E$ |
|---|---|---|---|---|---|---|
| [0231:015] | - | 742 | FALSE | 0.04 | YES | **YES** |
| [0331:033] | A | 0 | TRUE | 4.3 | NO | **NO** |
| [0331:033] | B | 0 | FALSE | 1000 | NO | |

Table 7: Final binary classification of the case study sample sentences

Since neither readings of [0331:033] are classified as such, the sentence is not an *it*-extraposition construct.

## 5. Evaluation

In order to provide a comprehensive picture of the system's performance, a twofold assessment is used. In the first evaluation, the system is exposed to the same sentence collection that assisted its development. Accordingly, results obtained from this evaluation reflect, to a certain degree, the system's optimal performance. The second evaluation aims at revealing the system's performance on unfamiliar texts by running the developed system on a random dataset drawn from the rest of the corpus. Two additional experiments are also conducted to provide an estimation of the system's performance over the whole corpus.

Three performance measures are used throughout the section: precision, recall, and the balanced F-measure (van Rijsbergen, 1979). Precision is defined as the ratio of correctly classified instances in a specific category (or a collection of categories) to the number of instances identified by the system as belonging to the category (categories). In other words, precision is calculated as $P = \frac{TP}{TP+FP}$, where $TP$ and $FP$ are the number of true positives and false positives respectively. Recall is defined as the ratio of correctly classified instances in a specific category (or a collection of categories) to the total number of instances in the category (categories), or $R = \frac{TP}{TP+FN}$, where $FN$ denotes the number of false negatives. Finally, the F-measure is the weighted harmonic mean of precision and recall used to indicate a system's overall performance. When precision and recall are weighted equally, as used in this study, the balanced F-measure is defined as $F = \frac{2PR}{P+R}$.

Following Efron and Tibshirani's (1993) Bootstrap method, 95% confidence intervals are obtained using the $2.5^{th}$ and $97.5^{th}$ percentiles of the bootstrap replicates and are provided alongside the system performance figures to indicate their reliability. The number of replicates is arbitrarily set at $B = 9999$, which is much greater than the commonly suggested value of 1000 (e.g., see Davison & Hinkley, 1997; Efron & Tibshirani, 1993) because pleonastic instances are sparse. In the case that a precision or recall value is 100%, the bootstrap percentile method reports an interval of 100%-100%, which makes little sense. Therefore, in this situation the adjusted Wald interval (Agresti & Coull, 1998) is presented instead. When two systems are compared, an approximate randomization test (Noreen, 1989) similar to that used by Chinchor (1992) is performed to determine if the difference is of statistical significance. The significance level $\alpha = 0.05$ and number of shuffles $R = 9999$, both chosen arbitrarily, are used where significance tests are performed.

### 5.1 Development Dataset

For the purpose of this study, the first 1000 occurrences of *it* from the WSJ corpus have been manually annotated by the authors[20]. A part of the set has also been inspected in order to determine the values of the constants specified in Section 3.5, and to develop the surface structure processor. The annotation process is facilitated by a custom-designed utility that displays each sentence within its context represented by a nine-sentence window containing the six immediately preceding sentences, the original, and the two sentences that follow. Post-annotation review indicates that this presentation of corpus sentences worked well. Except for a few (less than 0.5%) cases, the authors found no need to resort to broader contexts to understand a sentence; and under no circumstances were valid antecedents located outside the context window while no antecedent was found within it.

| **Category** | **Instances** | **Percentage** |
|---|---:|---:|
| Nominal Antecedent | 756 | 75.60% |
| Clause Antecedent | 60 | 6.00% |
| Extraposition | 118 | 11.80% |
| Cleft | 13 | 1.30% |
| Weather/Time | 9 | 0.90% |
| Idiom | 18 | 1.80% |
| Other | 26 | 2.60% |
| **Grand Total** | **1000** | **100.00%** |

Table 8: Profile of the development dataset according to the authors' annotation

Table 8 summarizes the distribution of instances in the dataset according to the authors' consensus. The category labeled 'Other' consists mostly of instances that do not fit well into any other categories, e.g. when the identified nominal antecedent is in plural or the antecedent is inferred, as well as certain confusing instances. Out of the twenty-six instances, only two might be remotely recognized as one of the types that interests this study:

[0101:007] And though the size of the loan guarantees approved yesterday is significant, recent experience with a similar program in Central America indicates that *it* could take several years before the new Polish government can fully use the aid effectively.

[0296:048] *It*'s just comic when they try to pretend they're still the master race.

Neither instance can be identified as anaphoric. However, the first construct has neither a valid non-extraposition version nor a valid *what*-cleft version, making it difficult to justify as an extraposition, while the *it* in the second case is considered to refer to the atmosphere aroused by the action detailed in the when-clause.

In order to assess whether the pleonastic categories are well-defined and the ability of ordinary language users to identify pleonastic instances, two volunteers, both native English speakers, are invited to classify the *it* instances in the development dataset. To help them concentrate on the pleonastic categories, the volunteers are only required to assign each instance to one of the following categories: referential, extraposition, cleft, weather/time, and idiom. The referential category covers instances with both nominal antecedents and clause antecedents, as well as instances with inferrable antecedents. Table 9 outlines both annotators' performance in reference to the authors' consensus. The degree of agreement between the annotators, measured by the kappa coefficient ($\kappa$; Cohen, 1960), is also given in the same table.

[20] Annotations are published as an online appendix at http://www.ece.ualberta.ca/~musilek/pleo.zip.

| **Category** | **Volunteer 1** | | | **Volunteer 2** | | | $\kappa$[a] |
|---|---|---|---|---|---|---|---|
| | **Precision** | **Recall** | **F-measure** | **Precision** | **Recall** | **F-measure** | |
| Referential | 99.38% | 95.49% | 97.40% | 96.38% | 98.10% | 97.23% | .749 |
| Extraposition | 82.54% | 88.14% | 85.25% | 88.68% | 79.66% | 83.93% | .795 |
| Cleft | 38.46% | 76.92% | 51.28% | 72.73% | 61.54% | 66.67% | .369 |
| Weather/Time | 66.67% | 44.44% | 53.33% | 75.00% | 33.33% | 46.15% | -.005 |
| Idiom | 39.39% | 72.22% | 50.98% | 50.00% | 61.11% | 55.00% | .458 |
| **Overall Accuracy/$\kappa$** | | **93.50%** | | | **94.20%** | | **.702** |

[a] Except for the Weather/Time category ($p = 0.5619$), all $\kappa$ values are statistically significant at $p < 0.0001$.

Table 9: Performance of the volunteer annotators on the development dataset (evaluated using the authors' annotation as reference) and the degree of inter-annotator agreement measured by Cohen's kappa ($\kappa$). The authors' annotations are refitted to the simplified annotation scheme used by the volunteers.

There are many factors contributing to the apparently low $\kappa$ values in Table 9, most notably the skewed distribution of the categories and inappropriate communication of the classification rules. As Di Eugenio and Glass (2004) and others pointed out, skewed distribution of the categories has a negative effect on the $\kappa$ value. Since the distribution of the *it* instances in the dataset is fairly unbalanced, the commonly-accepted guideline for interpreting $\kappa$ values ($\kappa > 0.67$ and $\kappa > 0.8$ as thresholds for tentative and definite conclusions respectively; Krippendorff, 1980) may not be directly applicable in this case. In addition, the classification rules are communicated to the annotators orally through examples and some of the not-so-common cases, such as the object *it*-extrapositions, might not have been well understood by both annotators. Another interesting note about the results is that there is a strong tendency for both annotators (albeit on different cases) to classify *it*-clefts as *it*-extrapositions. Rather than taking this as a sign that the cleft category is not well-defined, we believe it reflects the inherent difficulties in identifying instances pertaining to the category.

## 5.2 Baselines

Two baselines are available for comparison – the WSJ annotation, which is done manually and provided with the corpus; and the results from a replication of Paice and Husk's (1987) algorithm (PHA). It should be cautioned that, given the subjectivity of the issues discussed in this paper and lack of consensus on certain topics in the field of linguistics, recall ratios of the presented baseline results and the forthcoming results of the proposed system should not be compared quantitatively. For example, the original Paice and Husk algorithm does not recognize certain types of object extrapositions and does not always distinguish between individual types of pleonastic *it*; and the WSJ corpus has neither special annotation for parenthetical *it* (c.f. Section 3.2.1, Page 348, [0239:009]) nor an established annotation policy for certain types of object extrapositions (Bies, Ferguson, Katz, & MacIntyre, 1995). No attempts have been made to correct these issues.

Table 10 summarizes the performance of the baselines on the development dataset. As expected, Paice and Husk's (1987) algorithm does not perform very well since the WSJ articles are very different from, and tend to be more sophisticated than, the technical essays that the algorithm was designed for. Compared to the originally reported precision of 93% and recall of 96%, the replicated PHA yields only 54% and 75% respectively on the development dataset. The performance of the replica is largely in line with what Boyd et al. (2005) obtained from their implementation of the same algorithm on a different dataset.

| | WSJ Annotation | | Replicated PHA |
|---|---|---|---|
| **Measurement** | **Extraposition** | **Cleft** | **Overall**[a] |
| **Reference** | 118 | 13 | 140 |
| **Identified by Baseline** | 88 | 12 | 194 |
| **Baseline True Positives** | 87[b] | 12 | 105 |
| **Precision** | 98.86% | 100% | 54.12% |
| **Recall** | 73.73% | 92.31% | 75.00% |
| **F-measure** | 84.47% | 96.00% | 62.87% |

[a]Includes clefts, extrapositions, and time/weather cases.
[b]Based on manual inspection, two cases originally annotated as extrapositional in WSJ are determined as inappropriate. See discussions below.

Table 10: Performance of the baselines on the development dataset, evaluated against the authors' annotation.

The 31 (118 − 87) extrapositional cases that are not annotated in WSJ can be broken down into the following categories followed by their respective number of instances:

| **Category** | **Items** |
|---|---|
| Unrecognized | 3 |
| Object without complement | 1 |
| Parenthetical | 2 |
| Inappropriate non-extraposition | 18 |
| Agentless passive | 9 |
| *it seems/appears* ... | 4 |
| *it be worth* ... | 2 |
| Others | 3 |
| Valid non-extraposition | 10 |
| *too* ... *to* | 2 |
| Others | 8 |
| **Total** | **31** |

Table 11: Profile of the false negatives in the WSJ annotation in reference to the authors' annotation

By stating that the 'Characteristic of *it* extraposition is that the final clause can replace it', Bies et al. (1995) define the class in the narrowest sense. Since interpretation of the definition is entirely a subjective matter, there is no way of determining the real coverage of the annotations. However, from the portions of the corpus that have been reviewed, the practice of annotation is not entirely consistent.

Two sentences are marked as extraposition in the corpus but the annotators' consensus indicates otherwise. Considering the 'golden standard' status of the WSJ corpus, they are also listed here:

[0277:040] Moreover, as a member of the Mitsubishi group, which is headed by one of Japan's largest banks, *it* is sure to win a favorable loan.

[0303:006] *It* is compromises such as this that convince Washington's liberals that if they simply stay the course, this administration will stray from its own course on this and other issues.

The first sentence is considered dubious and most likely referring to the company that is a member of the Mitsubishi group. The second one is considered a cleft and is actually also marked as cleft in the corpus. Since it is the only case in the corpus with both annotations, the extraposition marking was considered a mistake and was manually removed.

The Paice and Husk (1987) algorithm suffers from false-positive *it* ... *that* and *it* ... *to* construct detection, which may be fixed by incorporating part-of-speech and phrase structure information together with additional rules. However, such fixes will greatly complicate the original system.

### 5.3 Results

On the development dataset, results produced by the proposed system are as follows:

| Measurement | Extraposition | Cleft | Weather/Time | Overall[a] |
|---|---|---|---|---|
| **Reference** | 118 | 13 | 9 | 140 |
| **Identified** | 116 | 13 | 10 | 139 |
| **True Positives** | 113 | 13 | 9 | 136 |
| **Precision** | 97.41% | 100.00% | 90.00% | 97.84% |
| **95% C.I.**[b] | 94.07-100.00% | 79.74-100.00% | 66.67-100.00% | 95.21-100.00% |
| **Recall** | 95.76% | 100.00% | 100.00% | 97.14% |
| **95% C.I.**[b] | 91.79-99.12% | 79.74-100.00% | 73.07-100.00% | 93.98-99.34% |
| **F-measure** | 96.58% | 100.00% | 94.74% | 97.49% |
| **95% C.I.** | 93.98-98.72% | - | 80.00-100.00% | 95.45-99.21% |

[a]Combining extraposition, cleft, and weather/time into one category.
[b]Adjusted Wald intervals are reported for extreme measurements.

Table 12: Performance of the system on the development dataset, evaluated using the authors' annotation as reference.

Further statistical significance tests reveal more information regarding the system's performance in comparison to that of the two volunteers and the baselines:

- Compared to both volunteer annotators, the system's better performance in all three pleonastic categories is statistically significant.
- In the extraposition category, the difference between the WSJ annotation's (higher) precision and that of the system is not statistically significant.
- Compared to Paice and Husk's (1987) algorithm, the system's higher precision is statistically significant.

| **Target System** | **Extraposition** | **Cleft** | **Weather/Time** |
|---|---|---|---|
| **Volunteer 1** | F-measure$^+$/$p < .001$ | F-measure$^+$/$p < .001$ | F-measure$^+$/$p = .033$ |
| **Volunteer 2** | F-measure$^+$/$p < .001$ | F-measure$^+$/$p = .007$ | F-measure$^+$/$p = .025$ |
| **WSJ Annotation** | Precision$^-$/$p = .630$ | F-measure$^+$/$p = 1.00$ | |
| **Replicated PHA** | (All Categories) Precision$^+$/$p < .001$ | | |

Table 13: Results of the statistical significance tests presented in the format Test Statistic$^{sign}$/$p$-value. A plus sign ($^+$) indicates that our system performs better on the reported measurement; otherwise a minus sign ($^-$) is used. If fair comparisons can be made for both precision and recall, the F-measure is used as the test statistic; otherwise the applicable measurement is reported.

Using the authors' annotation as reference, the system outperforms both human volunteers. While higher performance is usually desirable, in this particular case, it could indicate possible problems in the design of the experiment. Since the English language is not only used by its speakers but also shaped by the same group of people, it is impractical to have a system that 'speaks better English' than its human counterparts do. One plausible clue to the paradox is that an analytic approach is needed to gain insight into the issue of pronoun classification, but the casual English speakers do not see it from that perspective. As Green and Hecht (1992) and many others indicated, capable users of a language do not necessarily have the ability to formulate linguistic rules. However, these kinds of analytic skills is a prerequisite in order to explicitly classify a pronoun into one of the many categories. Thus, the true performance of casual speakers can only be measured by their ability to comprehend or produce the various pleonastic constructs. In addition, other factors, such as time constraints and imperfections in how the category definitions are conveyed, may also play a role in limiting the volunteers' performance. The authors' annotation, on the other hand, is much less influenced by such issues and is therefore considered expert opinion in this experiment. As shown in Section 5.2, the WSJ annotation of extrapositions and clefts, which is also considered expert opinion, is highly compatible with that of the authors. The differences between the two annotations can mostly be attributed to the narrower definition of extraposition adopted by the WSJ annotators. Therefore, the WSJ annotation's precision of 98.86% for extrapositions (when verified against the authors' annotation) is probably a more appropriate hint of the upper-limit for practically important system performance.

In the extraposition category, 279 individual cases passed the syntactic filters and were evaluated by search engine queries. Results of queries are obtained from Google through its web service, the Google SOAP[21] Search API. All three (116 − 113) cases of false-positives are caused by missing-object constructions and can be corrected using the patterns detailed in Section 3.4.3.

The five (118 − 113) false-negative cases are listed below:

[0283:013] The newspaper said *it* is past time for the Soviet Union to create unemployment insurance and retraining programs like those of the West.

[0209:040] "*It*'s one thing to say you can sterilize, and another to then successfully pollinate the plant," he said.

[0198:011] Sen. Kennedy said ... but that *it* would be a "reckless course of action" for President Bush to claim the authority without congressional approval.

[0290:049] Worse, *it* remained to a well-meaning but naive president of the United States to administer the final infamy upon those who fought and died in Vietnam.

[0085:047] "*It*'s not easy to roll out something that comprehensive, and make it pay," Mr. Jacob says.

Sentence [0283:013] is misplaced as weather/time. Sentence [0209:040] is not properly handled by the syntactic processing subcomponent. Sentences [0198:011] and [0290:049] involve complex noun phrases (underlined) at the object position of the matrix verbs – it is very difficult to reduce them to something more generic, such as the head noun only or a pronoun, and still remain confident that the original semantics are maintained. The last case, sentence [0085:047], fails because the full queries (containing part of the subordinate clause) failed to yield enough results and the stepped-down versions are overwhelmed by noise.

The last four false-negatives are annotated correctly in the WSJ corpus. The system's recall ratio on the 87 verified WSJ extraposition annotations is therefore 95.40%, comparable to the overall recall.

### 5.4 System Performance on Parser Output

Thus far, the system has been evaluated based on the assumption that the underlying sentences are tagged and parsed with (almost) perfect accuracy. Much effort has been made to reduce such dependency. For example, tracing information and function tags in the original phrase structures are deliberately discarded; and the system also tries to search for possible extraposed or cleft clauses that are marked as complements to the matrix object. However, deficiencies in tagging and parsing may still impact the system's performance. Occasionally, even the 'golden standard' manual markups appear problematic and happen to get in the way of the task.

[21] The Simple Object Access Protocol is an XML-based message protocol for web services.

It is therefore necessary to evaluate the system on sentences that are automatically tagged and parsed in order to answer the question of how well it would perform in the real world. Two state-of-the-art parsers are employed for this study: the reranking parser by Charniak and Johnson (2005), and the Berkeley parser by Petrov, Barrett, Thibaux, and Klein (2006). The system's performance on their respective interpretations of the development dataset sentences are reported in Tables 14 and 15. Table 16 further compares the system's real-world performance to the various baselines.

| Measurement | Extraposition | Cleft | Weather/Time | Overall[a] |
|---|---|---|---|---|
| **Reference** | 118 | 13 | 9 | 140 |
| **Identified** | 114 | 12 | 10 | 136 |
| **True Positives** | 110 | 12 | 9 | 132 |
| **Precision** | 96.49% | 100.00% | 90.00% | 97.06% |
| **95% C.I.**[b] | 92.68-99.20% | 78.40-100.00% | 66.67-100.00% | 93.92-99.32% |
| **Recall** | 93.22% | 92.31% | 100.00% | 94.29% |
| **95% C.I.**[b] | 88.43-97.41% | 73.33-100.00% | 73.07-100.00% | 90.18-97.81% |
| **F-measure** | 94.83% | 96.00% | 94.74% | 95.65% |
| **95% C.I.** | 91.60-97.49% | 84.62-100.00% | 80.00-100.00% | 93.08-97.90% |

[a]Combining extraposition, cleft, and weather/time into one category.
[b]Adjusted Wald intervals are reported for extreme measurements.

Table 14: Performance of the system on the development dataset parsed by the Charniak parser, using the authors' annotation as reference.

| Measurement | Extraposition | Cleft | Weather/Time | Overall[a] |
|---|---|---|---|---|
| **Reference** | 118 | 13 | 9 | 140 |
| **Identified** | 114 | 11 | 9 | 134 |
| **True Positives** | 111 | 10 | 8 | 130 |
| **Precision** | 97.37% | 90.91% | 88.89% | 97.01% |
| **95% C.I.** | 94.07-100.00% | 70.00-100.00% | 62.50-100.00% | 93.81-99.32% |
| **Recall** | 94.07% | 76.92% | 88.89% | 92.86% |
| **95% C.I.** | 89.47-98.18% | 50.00-100.00% | 62.50-100.00% | 88.44-96.91% |
| **F-measure** | 95.69% | 83.33% | 88.89% | 94.89% |
| **95% C.I.** | 92.75-98.17% | 62.50-96.55% | 66.67-100.00% | 92.02-97.35% |

[a]Combining extraposition, cleft, and weather/time into one category.

Table 15: Performance of the system on the development dataset parsed by the Berkeley parser, using the authors' annotation as reference.

**Comparing System Performance On Charniak Parser Output to:**

| Target System | Extraposition | Cleft | Weather/Time |
|---|---|---|---|
| **System w/o Parser** | F-measure$^-$/$p = .131$ | F-measure$^-$/$p = 1.00$ | F-measure$^=$/$p = 1.00$ |
| **Volunteer 1** | F-measure$^+$/$p = .001$ | F-measure$^+$/$p < .001$ | F-measure$^+$/$p = .030$ |
| **Volunteer 2** | F-measure$^+$/$p < .001$ | F-measure$^+$/$p = .041$ | F-measure$^+$/$p = .021$ |
| **WSJ Annotation** | Precision$^-$/$p = .368$ | F-measure$^=$/$p = 1.00$ | |
| **Replicated PHA** | (All Categories) Precision$^+$/$p < .001$ | | |

**Comparing System Performance On Berkeley Parser Output to:**

| Target System | Extraposition | Cleft | Weather/Time |
|---|---|---|---|
| **System w/o Parser** | F-measure$^-$/$p = .380$ | F-measure$^-$/$p = .128$ | F-measure$^-$/$p = 1.00$ |
| **Volunteer 1** | F-measure$^+$/$p < .001$ | F-measure$^+$/$p = .014$ | F-measure$^+$/$p = .061$ |
| **Volunteer 2** | F-measure$^+$/$p < .001$ | F-measure$^+$/$p = .314$ | F-measure$^+$/$p = .046$ |
| **WSJ Annotation** | Precision$^-$/$p = .627$ | F-measure$^-$/$p = .374$ | |
| **Replicated PHA** | (All Categories) Precision$^+$/$p < .001$ | | |

Table 16: Results of the statistical significance tests comparing the system's performance on parser output to that of various other systems, presented in the format Test Statistic$^{sign}$/$p$-value. A plus sign ($^+$) indicates that the proposed system performs better than the target system on the reported measurement; an equal sign ($^=$) indicates a tie; otherwise a minus sign ($^-$) is used. If fair comparisons can be made for both precision and recall, the F-measure is used as the test statistic; otherwise the applicable measurement is reported.

Further significance tests reveal that:

- using a parser has no statistically significant influence on the system's performance;
- the system outperforms both volunteer annotators in identifying *it*-extrapositions;
- regardless of the parser used, the difference between the system's performance and that of the WSJ annotation is not statistically significant; and
- regardless of the parser used, the system outperforms the Paice and Husk (1987) algorithm.

### 5.5 Correlation Analysis for Extrapositions

Figures 5 through 8 illustrate the correlation between the decision factors and the true expletiveness of the pronoun *it* in question. All 279 items that passed the initial syntactic filtering process are included in the dataset with the first 116 being extrapositional and the rest separated by a break on the X-axis. This arrangement is made in order to better visualize the contrast between the positive group and the negative group. In Figures 6 through 8, different grey levels are used to indicate the number of results returned by queries – the darker the shade, the more popular the construct in question is on the web. The constant $R_{exp} = 0.15$ is also indicated with a break on the Y-axis.

As illustrated, all factors identified in Section 3.5 are good indicators of expletiveness. $W$ (Figure 5) is the weakest of the four factors due to the number of false positives produced by incorrect language usage. This is clear evidence that the web is noisier than ordinary corpora and that the results counts from the web may not be appropriate as the sole decision-making factor. In comparison, $r$ (Figure 6) has almost perfect correlation with the expletiveness of instances. However, full versions of the queries usually return fewer results and in many cases yield too few results for expletive cases (unfilled items plotted on top of the graph indicate cases that do not have enough results, c.f. Section 3.5). The stepped-down versions of the queries (Figure 7), while being less accurate by themselves, serve well when used as 'back up', as illustrated by the $R$ plot (Figure 8). Part of the false-positive outliers on the $R$ plot are produced by full queries for expressions that are habitually associated with *it*, such as [0135:002] ' ... *said it <u>expects to post</u> sales in the current fiscal year* ... '. When used with a pronoun, these expressions usually describe information quoted from a person or organization already named earlier in the same sentence, making *it* a more natural choice of subject pronoun. Normally the problematic expressions take the form of `verb infinitive-complement`, i.e. $S$=TRUE. According to the decision process described in Section 3.5, $W$ is also considered in this situation, which effectively eliminates such noise.

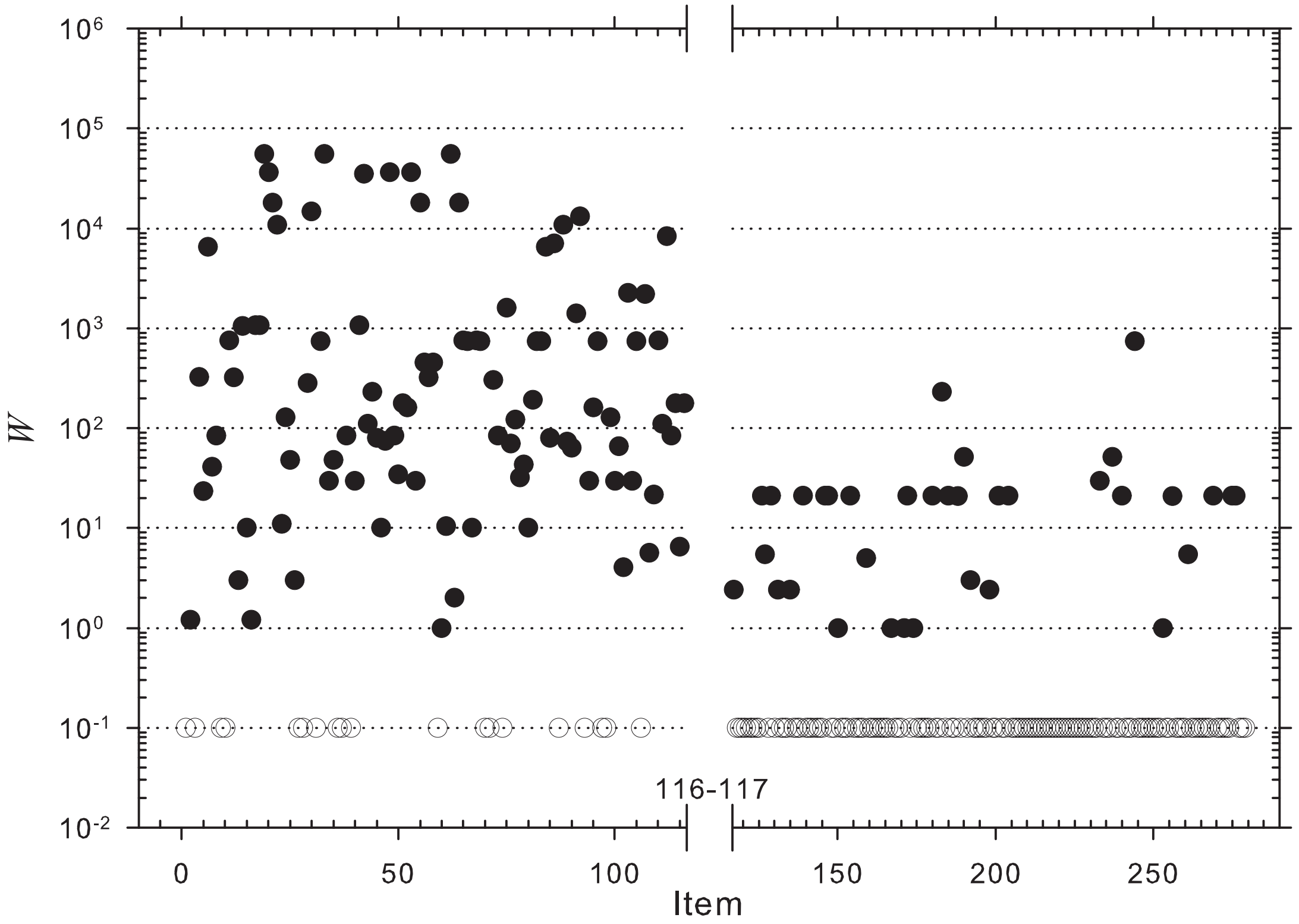

Figure 5: A scatter plot illustrating the correlation between $W$ (the estimated number of valid results returned by the *what*-cleft queries) and the expletiveness of the *it* instance. The extrapositional instances are arranged on the left side of the plot and the rest of the cases are on the right. If a query returns no valid results, the corresponding item is shown as a hollow circle on the bottom of the plot.

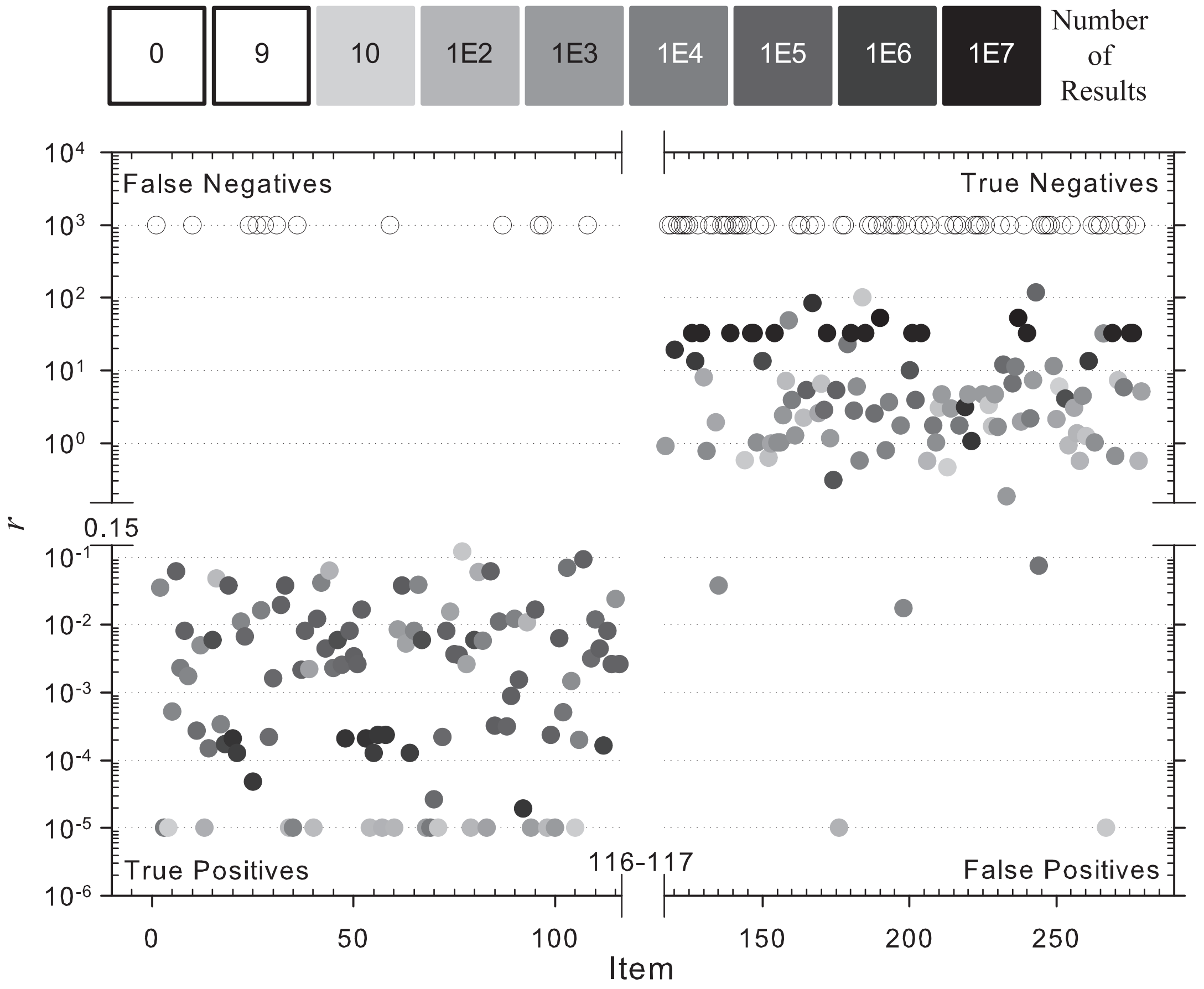

Figure 6: A scatter plot illustrating the correlation between $r$ (the ratio of the hit count produced by the expression with substitute pronouns to that of the original expression) and the expletiveness of the *it* instance. The extrapositional instances are arranged on the left side of the plot and the rest of the cases are to the right. The items are shaded according to the hit counts produced by the corresponding original expressions. If a query returns insufficient results, the corresponding item is shown as a hollow unshaded circle on the top of the plot.

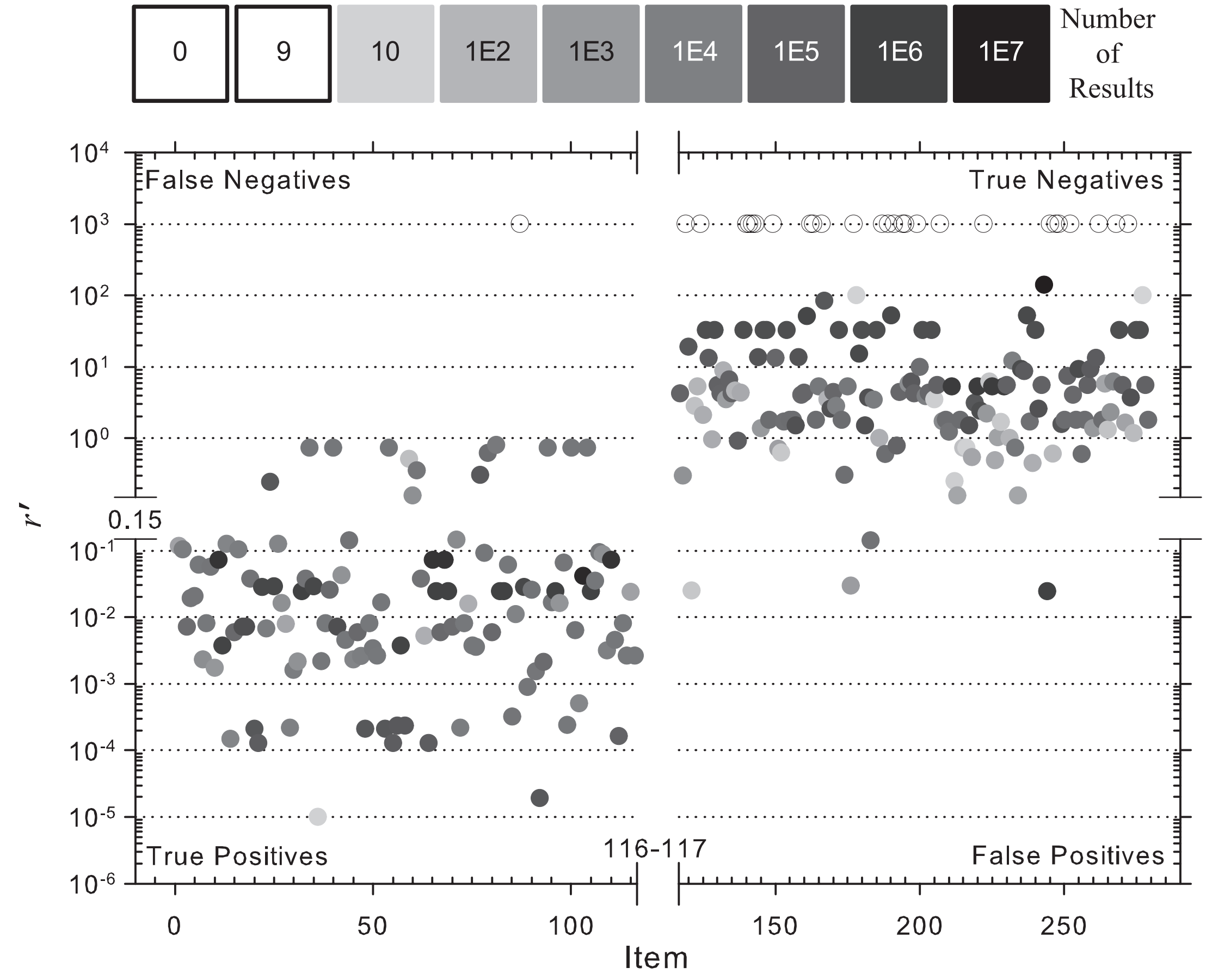

Figure 7: A scatter plot illustrating the correlation between $r'$ (similar to $r$ but for the stepped-down queries) and the expletiveness of the *it* instance. The extrapositional instances are arranged on the left side of the plot and the rest of the cases are to the right. The items are shaded according to the hit counts produced by the corresponding original expressions. If a query returns insufficient results, the corresponding item is shown as a hollow unshaded circle on the top of the plot.

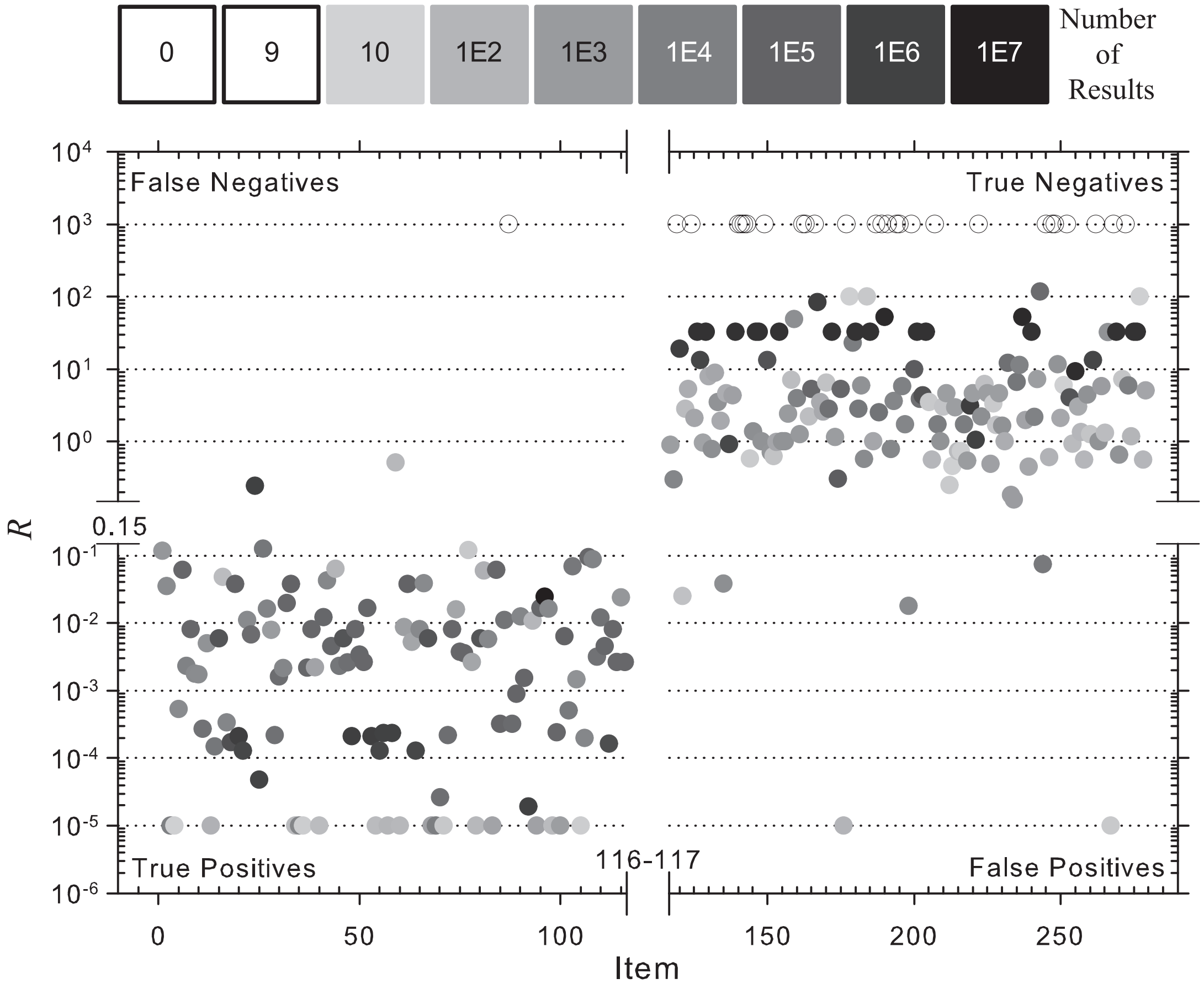

Figure 8: A scatter plot illustrating the correlation between $R$ (synthesized expletiveness; it takes the value of $r$ if the more complex queries produce enough results, and takes the value of $r'$ when they fail to do so) and the expletiveness of the *it* instance. The extrapositional instances are arranged on the left side of the plot and the rest of the cases are to the right. The items are shaded according to the hit counts produced by the corresponding original expressions. If a query returns insufficient results, the corresponding item is shown as a hollow unshaded circle on the top of the plot.

### 5.6 Generalization Study

In order to evaluate how well the system generalizes, 500 additional sample sentences are randomly selected from the rest of the WSJ corpus as the test dataset. The distribution of instances is comparable to that of the development dataset, as shown in Table 17.

| Category | Instances | Percentage |
|---|---|---|
| Nominal Antecedent | 375 | 75.00% |
| Clause Antecedent | 24 | 4.80% |
| Extraposition | 63 | 12.60% |
| Cleft | 8 | 1.60% |
| Weather/Time | 6 | 1.20% |
| Idiom | 11 | 2.20% |
| Other | 13 | 2.60% |
| **Grand Total** | **500** | **100.00%** |

Table 17: Profile of the test dataset according to the authors' annotation

As shown in Table 18, the overall level of inter-annotator agreement is slightly higher than that of the development dataset. Except for the idiom category, categorical $\kappa$ values are also higher than their counterparts on the development dataset. This discrepancy is most likely due to chance, since the two volunteers worked independently and started from different datasets (Volunteer 1 started from the development dataset and Volunteer 2 started from the test dataset).

| **Category** | **Volunteer 1** | | | **Volunteer 2** | | | $\kappa$[a] |
|---|---|---|---|---|---|---|---|
| | **Precision** | **Recall** | **F-measure** | **Precision** | **Recall** | **F-measure** | |
| Referential | 98.48% | 95.12% | 96.77% | 97.30% | 96.83% | 97.07% | .797 |
| Extraposition | 87.10% | 85.71% | 86.40% | 80.00% | 82.54% | 81.25% | .811 |
| Cleft | 29.41% | 62.50% | 40.00% | 57.14% | 50.00% | 53.33% | .490 |
| Weather/Time | 100.00% | 50.00% | 66.67% | 100.00% | 50.00% | 66.67% | .665 |
| Idiom | 31.82% | 53.85% | 40.00% | 47.06% | 61.54% | 53.33% | .280 |
| **Overall Accuracy/$\kappa$** | | **91.80%** | | | **92.80%** | | **.720** |

[a]All $\kappa$ values are statistically significant at $p < 0.0001$.

Table 18: Performance of the volunteer annotators on the test dataset (evaluated using the authors' annotation as reference) and the degree of inter-annotator agreement measured by Cohen's kappa ($\kappa$). The authors' annotations are refitted to the simplified annotation scheme used by the volunteers.

| | **WSJ Annotation** | | **Replicated PHA** |
|---|---|---|---|
| **Measurement** | **Extraposition** | **Cleft** | **Overall**[a] |
| **Reference** | 63 | 8 | 77 |
| **Identified by Baseline** | 54 | 6 | 97 |
| **Baseline True Positives** | 52 | 6 | 55 |
| **Precision** | 96.30% | 100.00% | 56.70% |
| **Recall** | 82.54% | 75.00% | 71.43% |
| **F-measure** | 88.89% | 85.71% | 63.22% |

[a]Includes clefts, extrapositions, and time/weather cases.

Table 19: Performance of the baselines on the test dataset, evaluated against the authors' annotation.

Table 19 summarizes the performance of the baselines on the test dataset. The two (54 − 52) false-positive extrapositions from the WSJ annotation are listed below together with their respective context:

> [1450:054-055] Another solution cities might consider is giving special priority to police patrols of small-business areas. For cities losing business to suburban shopping centers, *it* may be a wise business investment to help keep those jobs and sales taxes within city limits.

> [1996:061-062] You think you can go out and turn things around. *It*'s a tough thing when you can't.

The first case is considered referential, and the *it* in the second case is believed to refer to a hypothetical situation introduced by the *when*-clause.

#### 5.6.1 Performance Analysis

On the test dataset, the system is able to maintain its precision; it exhibits slight deterioration in recall but the overall performance is still within expectations. The findings are summarized in Table 20.

| Measurement | Extraposition | Cleft | Weather/Time | Overall[a] |
|---|---|---|---|---|
| **Reference** | 63 | 8 | 6 | 77 |
| **Identified** | 60 | 6 | 7 | 73 |
| **True Positives** | 58 | 6 | 6 | 70 |
| **Precision** | 96.67% | 100.00% | 85.71% | 95.89% |
| **95% C.I.**[b] | 91.38-100.00% | 64.26-100% | 50.00-100.00% | 90.77-100.00% |
| **Recall** | 92.06% | 75.00% | 100.00% | 90.91% |
| **95% C.I.**[b] | 84.85-98.25% | 40.00-100.00% | 64.26-100.00% | 84.15-97.01% |
| **F-measure** | 94.31% | 85.71% | 92.31% | 93.33% |
| **95% C.I.** | 89.60-98.11% | 57.14-100.00% | 66.67-100.00% | 88.75-97.10% |

[a]Combining extraposition, cleft, and weather/time into one category.
[b]Adjusted Wald intervals are reported for extreme measurements.

Table 20: Performance of the system on the test dataset, evaluated using the authors' annotation as reference.

149 instances were evaluated for extraposition using queries, covering 62 of the 63 extrapositions. The excluded case is introduced in the form of a direct question, whose particulars the syntactic processing subsystem is not prepared for. Of the other four false negatives, three involve noun phrases at the matrix object position. One of the two clefts that are not recognized arises out of imperfect processing in the corpus. In addition, the false positive in the weather/time category is caused by the verb '*hail*', which was treated as a noun by the system.

All five (63 − 58) false-negative extraposition cases are annotated in the corpus and the WSJ annotation agrees with the six clefts identified by the proposed system. Thus the system's recall ratio on the verified WSJ annotations is 90.38% for extraposition and 100% for cleft.

| Target System | Extraposition | Cleft | Weather/Time |
|---|---|---|---|
| **Volunteer 1** | F-measure$^+$/$p = .041$ | F-measure$^+$/$p = .005$ | F-measure$^+$/$p = .248$ |
| **Volunteer 2** | F-measure$^+$/$p = .002$ | F-measure$^+$/$p = .119$ | F-measure$^+$/$p = .254$ |
| **WSJ Annotation** | Precision$^-$/$p = .697$ | F-measure$^=$/$p = 1.00$ | |
| **Replicated PHA** | (All Categories) Precision$^+$/$p < .001$ | | |

Table 21: Results of the statistical significance tests, presented in the format Test Statistic$^{sign}$/$p$-value. A plus sign ($^+$) indicates that our system performs better on the reported measurement; an equal sign ($^=$) indicates a tie; otherwise a minus sign ($^-$) is used. If fair comparisons can be made for both precision and recall, the F-measure is used as the test statistic; otherwise the applicable measurement is reported.

**Performance on Charniak Parser Output**

| Measurement | Extraposition | Cleft | Weather/Time | Overall[a] |
|---|---|---|---|---|
| **Reference** | 63 | 8 | 6 | 77 |
| **Identified** | 58 | 7 | 7 | 72 |
| **True Positives** | 55 | 6 | 6 | 67 |
| **Precision** | 94.83% | 85.71% | 85.71% | 93.06% |
| **95% C.I.** | 88.24-100.00% | 50.00-100.00% | 50.00-100.00% | 86.36-98.51% |
| **Recall** | 87.30% | 75.00% | 100.00% | 87.01% |
| **95% C.I.**[b] | 78.26-95.08% | 37.50-100.00% | 64.26-100.00% | 78.95-94.12% |
| **F-measure** | 90.91% | 80.00% | 92.31% | 89.93% |
| **95% C.I.** | 84.75-95.77% | 50.00-100.00% | 66.67-100.00% | 84.30-94.57% |

**Performance on Berkeley Parser Output**

| Measurement | Extraposition | Cleft | Weather/Time | Overall[a] |
|---|---|---|---|---|
| **Reference** | 63 | 8 | 6 | 77 |
| **Identified** | 58 | 5 | 7 | 70 |
| **True Positives** | 56 | 5 | 6 | 67 |
| **Precision** | 96.55% | 100.00% | 85.71% | 95.71% |
| **95% C.I.**[b] | 91.11-100.00% | 59.90-100.00% | 50.00-100.00% | 90.28-100.00% |
| **Recall** | 88.89% | 62.50% | 100.00% | 87.01% |
| **95% C.I.**[b] | 80.60-96.23% | 25.00-100.00% | 64.26-100.00% | 79.22-93.90% |
| **F-measure** | 92.56% | 76.92% | 92.31% | 91.16% |
| **95% C.I.** | 87.14-96.97% | 40.00-100.00% | 66.67-100.00% | 85.94-95.52% |

[a] Combining extraposition, cleft, and weather/time into one category.
[b] Adjusted Wald intervals are reported for extreme measurements.

Table 22: Performance of the system on the test dataset using parser-generated output, evaluated using the authors' annotation as reference.

Results of the significance tests, summarized in Table 21, reveal the following additional information about the system's performance on the test dataset:

- the system's higher performance in recognizing *it*-extrapositions than both volunteers is statistically significant;
- in the extraposition category, the difference between WSJ annotation's (higher) precision and that of the system is not statistically significant; and
- the system outperforms the Paice and Husk (1987) algorithm, and the difference is statistically significant.

Tables 22 and 23 outline the system's performance on the test dataset when parsers are used. Again, both parsers cause slight deteriorations in system performance. However, such changes are not statistically significant. With either parser used, the system is able to perform as well as the WSJ annotations.

**Comparing System Performance On Charniak Parser Output to:**

| **Target System** | **Extraposition** | **Cleft** | **Weather/Time** |
|---|---|---|---|
| **System w/o Parser** | F-measure$^-/p = .125$ | F-measure$^-/p = 1.00$ | F-measure$^=/p = 1.00$ |
| **Volunteer 1** | F-measure$^+/p = .298$ | F-measure$^+/p = .013$ | F-measure$^+/p = .247$ |
| **Volunteer 2** | F-measure$^+/p = .022$ | F-measure$^+/p = .269$ | F-measure$^+/p = .246$ |
| **WSJ Annotation** | Precision$^-/p = .886$ | F-measure$^-/p = 1.00$ | |
| **Replicated PHA** | (All Categories) Precision$^+/p < .001$ | | |

**Comparing System Performance On Berkeley Parser Output to:**

| **Target System** | **Extraposition** | **Cleft** | **Weather/Time** |
|---|---|---|---|
| **System w/o Parser** | F-measure$^-/p = .501$ | F-measure$^-/p = 1.00$ | F-measure$^=/p = 1.00$ |
| **Volunteer 1** | F-measure$^+/p = .131$ | F-measure$^+/p = .035$ | F-measure$^+/p = .256$ |
| **Volunteer 2** | F-measure$^+/p = .009$ | F-measure$^+/p = .308$ | F-measure$^+/p = .27$ |
| **WSJ Annotation** | Precision$^-/p = .809$ | F-measure$^-/p = 1.00$ | |
| **Replicated PHA** | (All Categories) Precision$^+/p < .001$ | | |

Table 23: Results of the statistical significance tests comparing the system's performance on parser output to that of various other systems, presented in the format Test Statistic$^{sign}$/$p$-value. A plus sign ($^+$) indicates that the source system performs better on the reported measurement; an equal sign ($^=$) indicates a tie; otherwise a minus sign ($^-$) is used. If fair comparisons can be made for both precision and recall, the F-measure is used as the test statistic; otherwise the applicable measurement is reported.

### 5.6.2 Estimated System Performance on the Whole Corpus

The relative sparseness of clefts makes it hard to assess the real effectiveness of the proposed approach. To compensate for this, an approximate study is conducted. First, *it* instances in the whole corpus are processed automatically using the proposed approach. The identified cleft instances are then merged with those that are already annotated in the corpus to form an evaluation dataset of 84 sentences, which is subsequently verified manually. 76 instances out of the 84 are considered to be valid cleft constructs by the authors. Respective performances of the proposed approach and the WSJ annotation are reported in Table 24; the differences are not statistically significant.

| System | Total | Identified | Common | Precision | Recall[a] | F-measure[a] |
|---|---|---|---|---|---|---|
| **WSJ** | 76 | 66 | 63 | 95.45% | 82.94% | 88.73% |
| | | | 95% C.I.: | 89.55-100.00% | 74.32-90.79% | 82.86-93.79% |
| **Proposed Approach** | 76 | 75 | 70 | 93.33% | 92.11% | 92.72% |
| | | | 95% C.I.: | 87.50-98.65% | 85.53-97.40% | 87.84-96.65% |

[a]The reported recall ratios and F-measures are for the synthetic dataset only and cannot be extended to the whole corpus.

Table 24: Estimated system performance on *it*-cleft identification over the entire corpus

Three of the false positives produced by the proposed approach are actually extrapositions[22], which is expected (c.f. Footnote 12, Page 351). Thus, in a binary classification of pleonastic *it*, items in the cleft category will have higher contributions to the overall precision than they do for their own category. Until the whole corpus is annotated, it is impossible to obtain precise recall figures of either the WSJ annotations or the proposed approach. However, since the rest of the corpus (other than the synthetic dataset) does not contain any true positives for either system and contains the same number of false-negatives for both systems, the proposed system will maintain a higher recall ratio than that of the WSJ annotations on the whole corpus.

A similar experiment is conducted for extrapositions using sentences that are already annotated in the corpus. All 656 annotated extrapositional *it* instances are manually verified and 637 (97.10%) of them turn out to be valid cases. The system produced queries for 623 instances and consequently recognized 575 of them, translating into 90.27% (95% C.I. 89.01-93.56%) recall ratio on the verified annotations. Given the fact that on both the development dataset and the test dataset the proposed system yields slightly higher recall on the whole dataset than it does on the subsets identified by WSJ annotations, its performance for extrapositions on the whole WSJ corpus is likely to remain above 90% in recall.

Similar to the situation in the test based on random cases, a large portion of false-positives are contributed by imperfect handling of both surface structures and noun phrases in the matrix object position, particularly in the form of *it takes/took . . . to . . .* From additional experiments, it seems that this particular construct can be addressed with a different pattern, `what/whatever` *`it takes`* `to verb`, which eliminates the noun phrase. Alternatively, the construct could possibly be assumed as extrapositional without issuing queries at all.

[22]This kind of cleft can be separated from extrapositions using an additional pattern that attaches the prepositional phrase to the subordinate verb. However, the number of samples are too few to justify its inclusion in the study.

## 6. Discussion

In this paper a novel pleonastic-*it* identification system is proposed. Unlike its precursors, the system classifies extrapositions by submitting queries to the web and analyzing returned results. A set of rules are also proposed for classification of clefts, whose particular manner of composition makes it more difficult to apply the web-based approach. Components of the proposed system are simple and their effectiveness should be independent of the type of text being processed. As shown in the generalization tests, the system maintains its precision while recall degrades by only a small margin when confronted with unfamiliar texts. This is an indication that the general principles behind the system are not over-fitted to the text from which they were derived. Overall, when evaluated on WSJ news articles – which can be considered a 'difficult' type of nonfiction – the system is capable of producing results that are on par with or only slightly inferior to that of casually trained humans.

The system's success has important implications beyond the particular problem of pleonastic-*it* identification. First, it shows that the web can be used to answer linguistic questions that are based upon more than just simplistic semantic relationships. Second, the comparative study is an effective means to get highly accurate results from the web despite the fact that it is noisier than the manually compiled corpora. In addition, the success of the simple guidelines used in identifying clefts may serve as evidence that a speaker's intention can be heavily reflected by the surface structures of her utterance, in a bid to make it distinguishable from similarly constructed sentences.

Some problems are left unaddressed in the current study, most notably the handling of complex noun phrases and prepositional phrases. Generally speaking, its approach to query instantiation is somewhat crude. To solve the noun-phrase issue, a finer-grained query downgrading is proposed, viz. first to supply the query with the original noun phrase, then the head noun, and finally the adjective that modifies the head noun, if there is one. The effectiveness of this approach is to be determined. As discussed in Section 5.6.2, a special rule can be used for the verb *take*. This, however, may open the door to exception-based processing, which contradicts the principle of the system to provide a unified approach to pleonastic pronoun identification. Overall, much more data and further experiments are needed before the query instantiation procedures can be finalized.

Aside from the two sets of patterns that are currently in use, other information can be used to assess the validity of a possible extraposition. For example, in extrapositions the matrix verbs are much more likely to remain in present tense than past tense, the noun phrases (if any) at the matrix object position are more likely to be indefinite, and the extraposed clauses are generally longer than the matrix verb phrases. A fuzzy-based decision system with multiple input variables could possibly provide significant performance gains.

Although the system is able to yield reasonable performances on the output of either parser tested, both of them introduce additional errors to the final results. On the combined dataset of development and test items, both parsers cause statistically significant deteriorations in performance at a significance level of 0.1 (Charniak parser: $p=0.008$ for F-measure on extrapositions; $p=0.071$ for F-measure on clefts). It is possible that incorporating a pattern-based method will compensate for the problems caused by imperfect parsing and further improve recall ratios; however, more data is needed to confirm this.

Another concern is that the syntactic processing component used in the system is limited. This limitation, caused by the designer's lack of exposure to a large variety of different constructs, is essentially different from the problem imposed by the limited number of patterns in some previous systems. Eventually, for the proposed system, this limitation can be eliminated. To illustrate, the current design is not able to correctly process sentences like *what difference does it make which I buy*; however, it only takes minor effort to correct this by upgrading the subsystem so that it recognizes pre-posed objects. Each such upgrade, which may be performed manually or even automatically through some machine-learning approaches, solves one or more syntactic problems and moves the system closer to being able to recognize all grammatically valid constructs. In contrast, it will take considerably more effort to patch the rigidly defined rules or to upgrade the word lists before the rule-based systems can achieve comparable performances.

During the writing of this article, Google deprecated their SOAP-based search API. This move makes it technically difficult to precisely replicate the results reported in this study since other search engines lack the ability to process alternate expressions (i.e. $\text{Word}_A$ OR $\text{Word}_B$) embedded within a quoted query. To use a different search engine, the matrix verbs should not be expanded but should instead be converted to their respective third-person singular present form only. Stubs should also be in their simplest form only, as described in earlier sections. From preliminary experiments it also seems possible to replace the combination of *which/who/this/he* with *they* alone, plus some necessary changes to maintain number agreement among the constituents of the queries. These changes may have some negative effects on the final outcome of the system, but they are unlikely to be severe.

Like most other NLP tasks, classifying the usage of *it* is inherently difficult, even for human annotators who already have some knowledge about the problem – it is one thing to speak the language, and another to then clearly explain the rationale behind a specific construct. Although it is widely accepted that an extrapositional *it* is expletive, the line between extrapositional cases and referential ones can sometimes be very thin. This is clearly manifested by the existence of truncated extrapositions (Gundel et al., 2005), which obviously have valid referential readings. Similar things can be said about the relationship among all three pleonastic categories as well as idioms. For example, Paice and Husk classify '*it remains to* ... ' as an idiom while the same construct is classified as an extraposition in our evaluations. Aside from applying the syntactic guidelines proposed in this study, it is assumed during the annotation process that an extraposition should have either a valid non-extraposed reading or a valid what-cleft reading. It is also assumed that a cleft should generate a valid non-clefted reading by joining the clefted constituent directly to the cleft clause without any leading relative pronoun or adverb. In light of the subjective nature of the problem, our annotations are published on the web as an online appendix to better serve readers.